\documentclass[10pt,twocolumn,letterpaper]{article}

\usepackage{iccv}              
\usepackage{multirow}
\usepackage{colortbl}
\definecolor{lightred}{rgb}{1, 0.6, 0.6}
\definecolor{lightorange}{rgb}{1, 0.8, 0.5}
\definecolor{lightyellow}{rgb}{1, 1, 0.6}
\usepackage{bm}
\usepackage{rotating} 
\usepackage{stfloats}
\usepackage{subcaption}
\usepackage{graphicx}
\usepackage[accsupp]{axessibility}  

\definecolor{iccvblue}{rgb}{0.21,0.49,0.74}
\usepackage[pagebackref,breaklinks,colorlinks,allcolors=iccvblue]{hyperref}

\def\paperID{7074} 
\def\confName{ICCV}
\def\confYear{2025}

\title{SEGS-SLAM: Structure-enhanced 3D Gaussian Splatting SLAM with Appearance Embedding}

\author{Tianci Wen$^{1,2}$ \quad 
Zhiang Liu$^{1,2}$ \quad 
Yongchun Fang$^{1,2\ *}$  \\
$^{1}$ IRAIS, tjKLIR, College of Artificial Intelligence, Nankai  University\\
$^{2}$ IITRS, Shenzhen Research Institute, Nankai University\\
}

\begin{document}
\twocolumn[{%
\renewcommand\twocolumn[1][]{#1}%
\maketitle
\begin{center}
  \centering
        \includegraphics[width=1\linewidth]{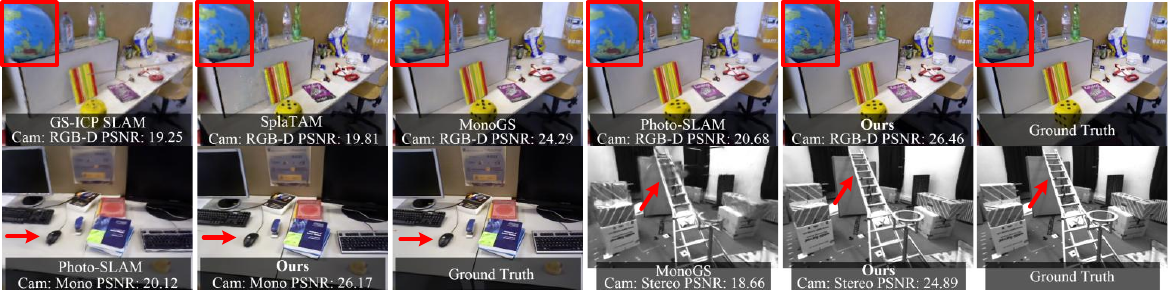}
   \captionof{figure}{Our method SEGS-SLAM outperforms SOTA methods  (GS-ICP SLAM \cite{GS-ICPSLAM2024}, Photo-SLAM \cite{Photo-SLAM2024}, SplaTAM \cite{SplaTAM2024}, MonoGS \cite{MonoGS2024}) in photorealistic mapping quality across monocular, stereo, and RGB-D cameras. The scenes are from TUM RGB-D dataset \cite{TUMRGB-D2012} (the top row and the left three images in the bottom row) and the EuRoC MAV dataset \cite{EuRoC2016} (the right three images in the bottom row).  Non-obvious differences in quality are highlighted by arrows/insets.  
   }   
   \label{fig:first_page}
\end{center}
}]
\renewcommand{\thefootnote}{}
\footnotetext{$^{*}$ Corresponding author.}
\begin{abstract}
3D Gaussian splatting (3D-GS) has recently revolutionized novel view synthesis in the simultaneous localization and mapping (SLAM) problem. However, most existing algorithms fail to fully capture the underlying structure, resulting in structural inconsistency. Additionally, they struggle with abrupt appearance variations, leading to inconsistent visual quality. To address these problems,  we propose SEGS-SLAM, a structure-enhanced 3D Gaussian Splatting SLAM, which achieves high-quality photorealistic mapping. Our main contributions are two-fold. First, we propose a structure-enhanced photorealistic mapping (SEPM) framework that, for the first time, leverages highly structured point cloud to initialize structured 3D Gaussians, leading to significant improvements in rendering quality. Second, we propose Appearance-from-Motion embedding (AfME), enabling 3D Gaussians to better model image appearance variations across different camera poses. Extensive experiments on monocular, stereo, and RGB-D datasets demonstrate that SEGS-SLAM significantly outperforms state-of-the-art (SOTA) methods in photorealistic mapping quality, e.g., an improvement of $19.86\%$ in PSNR over MonoGS on the TUM RGB-D dataset for monocular cameras. The project page is available at \url{https://segs-slam.github.io/}.
\end{abstract}

\section{Introduction}
\label{sec:intro}

Visual simultaneous localization and mapping (SLAM) is a fundamental problem in 3D computer vision, with wide applications in autonomous driving, robotics, virtual reality, and augmented reality. SLAM aims to construct dense or sparse maps to represent the scene. Recently, neural radiance fields (NeRF) \cite{NeRF2020} has been integrated into SLAM pipelines, significantly enhancing scene representation capabilities. The latest advancement in radiance field rendering is 3D Gaussian splatting (3D-GS) \cite{3DGS2023}, an explicit scene representation that achieves revolutionary improvements in rendering and training speed. Recent SLAM works \cite{Photo-SLAM2024, MonoGS2024, SplaTAM2024, GS-SLAM2024, RTG-SLAM2024, CG-SLAM2024, GS-ICPSLAM2024,Sgs-slam2024, DG-SLAM} incorporating 3D-GS have demonstrated that explicit representations provide more promising rendering performance when compared with implicit ones.



However, most SLAM algorithms based on 3D-GS have neglected the latent structure in the scene, which constrains their rendering quality. While some methods \cite{SplaTAM2024, GS-SLAM2024, RTG-SLAM2024, GS-ICPSLAM2024, CG-SLAM2024, MonoGS2024, Gaussian-SLAM2312, Sgs-slam2024,  DG-SLAM} have explored improvements in accuracy, efficiency, and semantics, insufficient structural exploitation remains a critical issue. For instance, as evidenced by the 2nd row in \cref{fig:first_page}, MonoGS \cite{MonoGS2024} produces a highly disorganized reconstruction of the ladder structure due to this limitation.  In contrast, few methods, like Photo-SLAM \cite{Photo-SLAM2024}, leverage scene structure. Photo-SLAM \cite{Photo-SLAM2024} initializes 3D Gaussians with point cloud obtained from indirect visual SLAM and incorporates a geometry-based densification module. In the original 3D-GS, 3D Gaussians are initialized from COLMAP \cite{colmap} points. Since indirect visual SLAM and COLMAP~\cite{colmap} share similar pipeline structures, the generated point clouds exhibit similar intrinsic properties.  Hence, the 3D Gaussians of Photo-SLAM \cite{Photo-SLAM2024} converge to a relatively optimal result with fewer iterations, yet it still underutilizes the underlying scene structure.  The blurry reconstruction of the mouse edge by Photo-SLAM \cite{Photo-SLAM2024} is still apparent as shown in the 2nd row of \cref{fig:first_page}.

Another partially unresolved challenge in these methods  \cite{SplaTAM2024, GS-SLAM2024, RTG-SLAM2024, GS-ICPSLAM2024, CG-SLAM2024, MonoGS2024, Photo-SLAM2024, Gaussian-SLAM2312, Sgs-slam2024, DG-SLAM} is the significant appearance variations within the scene (\eg, exposure, lighting). To address this issue, NeRF-W \cite{NeRF-W2021} refines the appearance embeddings (AE) in NRW \cite{NRW} and introduces them into NeRF.  However, AE has a notable limitation:  
its training involves each ground-truth image from the test set.
In novel view synthesis tasks, the test set contains 12.5\% of all views, whereas in SLAM tasks, this increases to 80\%, making it more challenging for AE to accurately predict appearance in novel views. Additionally, these approaches \cite{SplaTAM2024, GS-SLAM2024, RTG-SLAM2024, GS-ICPSLAM2024, CG-SLAM2024, MonoGS2024, Photo-SLAM2024, Scaffold-GS2024, Gaussian-SLAM2312, Sgs-slam2024, DG-SLAM} fail to capture high-frequency details (\eg, object edges, complex texture regions). FreGS \cite{FreGS2024} combines frequency regularization to model the local details, but its effectiveness is constrained by the use of a single-scale frequency spectrum.

To address the above limitations, this paper presents SEGS-SLAM, a novel 3D Gaussian Splatting SLAM system. First, we investigate the benefits of leveraging scene structure for improving rendering accuracy. While point cloud produced by ORB-SLAM3 \cite{ORB-SLAM32021} preserves strong latent structure, we observe that the anchor-based 3D Gaussians in \cite{Scaffold-GS2024} effectively leverage the underlying structure. Motivated by this, we propose a structure-enhanced photorealistic mapping (SEPM) framework, which initializes anchor points using ORB-SLAM3 \cite{ORB-SLAM32021} point cloud, significantly enhancing the utilization of scene structure. Experimental results validate the effectiveness of this simple yet powerful strategy, and we hope this insight will inspire further research in this direction. Second, we propose Appearance-from-Motion embedding (AfME), which takes poses as input and eliminates the need for training on the left half of each ground-truth image in the test set. We further introduce a frequency pyramid regularization (FPR) technique to better capture high-frequency details in the scene. The main contributions of this work are as follows: 
\begin{enumerate}
    \item To our knowledge, structure-enhanced photorealistic mapping (SEPM) is the first SLAM framework that intializes anchor points with ORB-SLAM3 point cloud to strengthen the utilization of scene structure, leading to significant rendering improvements. 
    \item We propose Appearance-from-Motion embedding (AfME), which models per-image appearance variations into a latent space extracted from camera pose. 
    \item Extensive evaluations on various public datasets demonstrate that our method significantly surpasses state-of-the-art (SOTA) methods in photorealistic mapping quality across monocular, stereo, and RGB-D cameras, while maintaining competitive tracking accuracy. 
\end{enumerate}


\section{Related Work}
\label{sec:Related}

\noindent\textbf{Visual SLAM.} Traditional visual SLAM methods can be classified into two categories: indirect methods and direct methods. Indirect methods \cite{PTAM, ORB-SLAM32021} rely on extracting and tracking features between consecutive frames to estimate poses and build sparse maps by minimizing a reprojection error, including ORB-SLAM3 \cite{ORB-SLAM32021}. Direct methods \cite{LSD-SLAM, DSO} estimate motion and structure by minimizing a photometric error, which can build sparse or semi-dense maps. Recently, some methods \cite{CodeSLAM, NodeSLAM, DROID-SLAM2021, DPVO, MS_SLAM_2024CVPR} have integrated deep learning into visual SLAM systems. Among them, the current SOTA method is Droid-SLAM \cite{DROID-SLAM2021}. More recently, Lipson \etal \cite{MS_SLAM_2024CVPR} combine optical flow prediction with a pose-solving layer to achieve camera tracking.  Our approach favors traditional indirect visual SLAM. 

\noindent\textbf{Implicit Representation based SLAM.} 
iMAP \cite{iMAP2021} pioneers the use of neural implicit representations to achieve tracking and mapping through reconstruction error. Subsequently, many works \cite{NICE-SLAM2022, Vox-Fusion2022, Orbeez-SLAM2023, ESLAM2023,Co-SLAM2023, SNI-SLAM2024, NGEL-SLAM2024,IBD-SLAM_2024_CVPR,NICER-SLAM2024} have explored new representation forms, including voxel-based neural implicit surface representation  \cite{Vox-Fusion2022}, multi-scale tri-planes \cite{ESLAM2023} and point-based neural implicit representation \cite{Point-SLAM2023}.  Recently, SNI-SLAM \cite{SNI-SLAM2024} and IBD-SLAM \cite{IBD-SLAM_2024_CVPR} introduce a hierarchical semantic representation and an xyz-map representation, respectively. Some works \cite{GO-SLAM2023, Loopy-SLAM2024, PLGSLAM_2024_CVPR, CP-SLAM2023, UncLe-SLAM2023, HI-SLAM2024_NeRF, NIS-SLAM2024}  address other challenges, including loop closure \cite{GO-SLAM2023, Loopy-SLAM2024}. However, most efforts focus on scene geometry reconstruction, with Point-SLAM \cite{Point-SLAM2023} extending to novel view synthesis. Moreover, the NeRF models used in these methods do not account for appearance variations.

\noindent\textbf{3D Gaussian Splatting based SLAM.} 
Recently, an explicit representation, 3D-GS \cite{3DGS2023}, is introduced into visual SLAM. Most methods enhance RGB-D SLAM in rendering quality \cite{GS-ICPSLAM2024, SplaTAM2024, GS-SLAM2024, Gaussian-SLAM2312}, efficiency \cite{RTG-SLAM2024, GS-ICPSLAM2024}, robustness \cite{CG-SLAM2024, DG-SLAM}, and semantics \cite{Sgs-slam2024}. For example, GS-ICP SLAM\cite{GS-ICPSLAM2024} achieves photorealistic mapping by fusing 3D-GS with Generalized ICP. Few methods improve SLAM rendering accuracy and efficiency across monocular, stereo, and RGB-D cameras, with MonoGS \cite{MonoGS2024} and Photo-SLAM \cite{Photo-SLAM2024} as exceptions. However, neither of these methods effectively leverages latent scene structures or enhances the modeling of scene details, which limits their rendering quality. To address this, our proposed SEGS-SLAM further reinforces the utilization of global scene structure.

\begin{figure*}[t]
  \centering
   \includegraphics[width=0.9\linewidth]{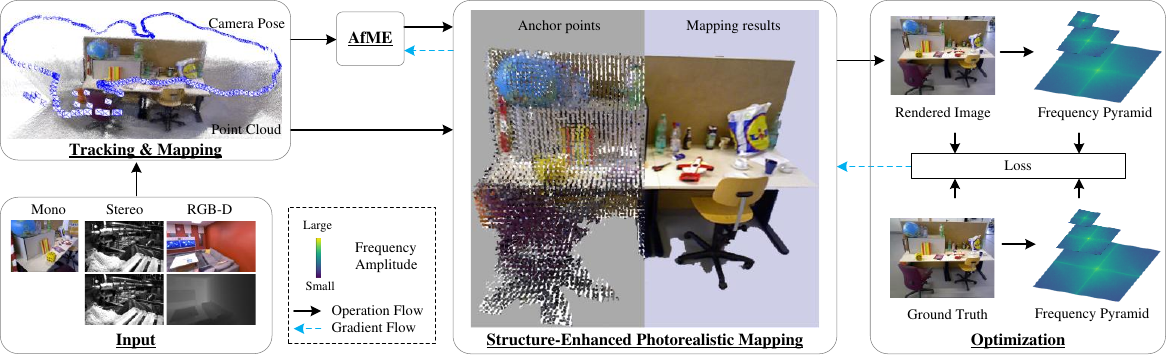}
   \caption{Overview of our method. Our method supports monocular, stereo, and RGB-D cameras. The input image stream is processed by the localization and geometric mapping modules, generating point cloud and accurate poses. SEPM incrementally initializes anchor points (middle) based on the point cloud (top left), which preserves the underlying structure. The poses are then fed into the AfME to model appearance variations in the scene. Additionally, we introduce FPR to improve the reconstruction of high-frequency details in the scene.} 
   \label{fig:diagram}
\end{figure*}
\label{sec:method}

\section{ Preliminaries}
In this section, we first introduce the structured 3D Gaussian splatting in \cite{Scaffold-GS2024}. Subsequently, we review the localization and mapping process of ORB-SLAM3 \cite{ORB-SLAM32021}.
\subsection{ Structured 3D Gaussian Splatting} \label{sec::scaffoldgs}
Structured 3D Gaussians is a hierarchical representation in \cite{Scaffold-GS2024}. They construct anchor points by voxelizing the point cloud obtained from COLMAP \cite{colmap}. An anchor point is the center $ \mathrm{\mathbf{t}}_v$ of a voxel, equipped with a context feature $\hat{f}_v \in \mathbb{R}^{32}$, a scale factor $l_v \in \mathbb{R}^{3}$, and a set of $k$ learnable offsets $\mathrm{\mathbf{O}}_v = \{\mathcal{O}_0,\ldots,\mathcal{O}_{k-1} \} \in \mathbb{R}^{k \times 3}$. For each anchor point $\mathrm{\mathbf{t}}_v$, they generate $k$ 3D Gaussians, whose positions are calculated as:
\begin{equation} \label{eq:pos}
\{\mu_0,\ldots,\mu_{k-1}\} = \mathrm{\mathbf{t}}_v + \{\mathcal{O}_0,\ldots,\mathcal{O}_{k-1} \} \cdot l_v.
\end{equation}
Other parameters of $k$ 3D Gaussians are decoded using individual MLPs, denoted as $M_\alpha$, $M_c$, $M_q$, and $M_s$, respectively. The colors of the Gaussians are obtained as follows:
\begin{equation} 
\{c_0,\dots,c_{k-1}\} = M_C(\hat{f}_v,\ \delta_{vc},\ \vec{\mathrm{\mathbf{d}}}_{vc}),
\label{eq:color_mlp1}
\end{equation}
where $\delta_{vc} = \| \mathrm{\mathbf{t}}_v-\mathrm{\mathbf{t}}_c\|_2$ is the relative distance between camera position $\mathrm{\mathbf{t}}_c$ and an anchor point and $\vec{\mathrm{\mathbf{d}}}_{vc} = (\mathrm{\mathbf{t}}_v-\mathrm{\mathbf{t}}_c) / \delta_{vc}$ is their viewing direction. The opacity $\{\alpha_i\}$, quaternion $\{q_i\}$, and scale $\{s_i\}$ are similarly obtained. After obtaining the parameters of each 3D Gaussian $G(x)$ within the view frustum, it is projected onto the image plane to form a 2D Gaussian $G^\prime_i(x^\prime)$. Following 3D-GS \cite{3DGS2023}, a tile-based rasterizer is used to sort the 2D Gaussians, and $\alpha-$blending is employed to complete the rendering:
\begin{equation} 
    C(x^\prime) = \sum_{i\in N} c_i \delta_i \prod_{j=1}^{i-1}(1 - \delta_j), \quad \delta_i = \alpha_i G^\prime_i(x^\prime),
  \label{eq:rendering}    
\end{equation}
where $x^\prime$ is the pixel position and $N$ is the number of corresponding 2D Gaussians for each pixel. 

\subsection{Localization and Geometry Mapping} \label{sec::LAGM}
ORB-SLAM3 \cite{ORB-SLAM32021} can track camera poses and generate point cloud accurately. The camera pose is represented as $(\mathrm{\mathbf{R}}, \mathrm{\mathbf{t}})$, where $\mathrm{\mathbf{R}}\in \mathrm{SO(3)}$ denotes orientation and $\mathrm{\mathbf{t}}\in \mathbb{R}^3$ represents position. The camera poses $(\mathrm{\mathbf{R}}, \mathrm{\mathbf{t}})$ and the point cloud $\{\mathrm{\mathbf{P}}_0,\dots,\mathrm{\mathbf{P}}_\eta \} \in \mathbb{R}^{\eta \times 3}$ of the scene can be solved through local or global bundle adjustment (BA):
\begin{align}
\{\mathrm{\mathbf{P}}_m,\mathrm{\mathbf{R}}_l,\mathrm{\mathbf{t}}_l\} &= \mathop{\mathrm{argmin}}\limits_{\mathrm{\mathbf{P}}_m,\mathrm{\mathbf{R}}_l,\mathrm{\mathbf{t}}_l} \sum_{\kappa\in \mathcal{K}_L\cup \mathcal{K}_F}  \sum_{j\in \mathcal{X}_k} \rho(E(\kappa,j)), \\
E(\kappa, j) &= \|\mathrm{\mathbf{p}}_j - \pi(\mathrm{\mathbf{R}}_\kappa\mathrm{\mathbf{P}}_j+\mathrm{\mathbf{t}}_\kappa) \|^2_{\Sigma_g},
\end{align}
where  $m\in {P}_L$, $l\in \mathcal{K}_L$, $\mathcal{K}_L$ is a set of covisible keyframes, ${P}_L$ are the points seen in $\mathcal{K}_L$, $\mathcal{K}_F$ are other keyframes, $\mathcal{X}_k$ is the set of matched points between a keyframe $\kappa$ and ${P}_L$, $\rho$ is the robust Huber cost function, $E(\kappa,j)$ is the reprojection error between the matched 3D points $\mathrm{\mathbf{P}}_j$ and 2D feature points $\mathrm{\mathbf{p}}_j$, $\pi$ is the projection function,  and $\Sigma_g$ denotes the covariance matrix associated with the keypoint's scale. We provide more details in \cref{sec:sup_SLAM} of supplementary material.  
\section{SEGS-SLAM}
In this section, we first give an overview of our system. We then present details of two key innovations: SEPM and AfME. Finally, we provide an introduction to our training loss and FPR. The overview of our SEGS-SLAM is summarized in \cref{fig:diagram}. First, the input image stream is processed through tracking and geometric mapping process in \cref{sec::LAGM}  to obtain camera poses and point cloud.  On one hand, SEPM voxelizes the point cloud to initialize anchors, enhancing the exploitation of underlying scene structure. On the other hand, AfME encodes the camera poses to model appearance variations. Finally, the rendered images are supervised by ground-truth images, with the assistance of FPR, while jointly optimizing the parameters of the 3D Gaussians and the weights of AfME throughout training.


 \begin{figure}[t]
	\centering
    \begin{subfigure}[ht]{1\linewidth}
    \includegraphics[width=1\linewidth]{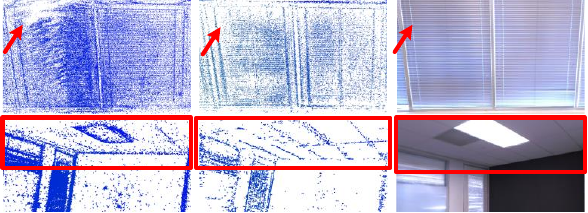}
    \end{subfigure}
    \begin{subfigure}[ht]{0.32\linewidth}
            \vspace{-0.15cm}
        \caption{ Photo-SLAM}
    \end{subfigure}   
    \begin{subfigure}[ht]{0.32\linewidth}
            \vspace{-0.15cm}
        \caption{Ours w SEPM only}
    \end{subfigure}    
    \begin{subfigure}[ht]{0.32\linewidth}
            \vspace{-0.15cm}
        \caption{  Ground Truth}
    \end{subfigure}
	\caption{Visualization of the Photo-SLAM's 3D Gaussians and of our method's anchor points using only SEPM  after 30k iterations. SEPM enhances the underlying structure of the 3D representation.  
   }
	\label{fig:structure}
\end{figure}

\subsection{Structure-Enhanced Photorealistic Mapping}
A common limitation of existing 3DGS-based SLAM systems is the gradual degradation of the underlying structure of Gaussians during optimization, limiting the quality of the rendered results. \textit{Our key insight is that preserving strong scene structure throughout the optimization process is crucial for achieving high-fidelity rendering.} Inspired by this, we make several observations. First, ORB-SLAM3 \cite{ORB-SLAM32021} generates a point cloud that preserves the scene structure well. Photo-SLAM \cite{Photo-SLAM2024} initializes 3D Gaussians with the point cloud, but as shown in the \cref{fig:structure} (a), relying solely on the latent structure of the point cloud is insufficient, as Gaussians still undergo structural degradation during optimization. We further observe that the hierarchical structure in \cite{Scaffold-GS2024} leverages scene structure, using anchor points to manage 3D Gaussians. These anchor points remain fixed during the optimization process, ensuring that 3D Gaussians retain the underlying structure of the scene.

Based on this observation, we propose incrementally voxelizing the point cloud $P_k$ of each keyframe to construct anchor points, as follows:  
\begin{equation} \label{eq:voxel}
V_k =  \{\lfloor \frac{P_k}{\epsilon} \rceil\}\cdot \epsilon,
\end{equation} 
where $V_k \in R^{N\times3}$ denotes voxel centers, $\epsilon$ is the voxel size, and $\{\cdot\}$ is the operation to remove redundant points. 

\cref{fig:diagram} visualizes this process. The ORB-SLAM3 point cloud (top left) is voxelized into the anchor points (middle), which preserve the underlying structure effectively. In this way, the structural prior from ORB-SLAM3's point cloud and the anchor-based organization are seamlessly fused, enhancing the exploitation of the underlying structure. Specifically, the anchor points inherit the strong structural properties of the point cloud and, due to their fixed positions, consistently preserve the latent scene structure throughout training. As shown in the \cref{fig:structure} (b), this effectively prevents structural degradation of Gaussians over training, a key issue in prior methods. This strategy yields substantial improvements in rendering accuracy, as demonstrated in our ablation studies. Although conceptually simple, our approach proves highly effective. 
After anchor points construction, 3D Gaussians are then generated according to \cref{eq:pos}, \cref{eq:color_mlp}, and rendered via \cref{eq:rendering}. 

\subsection{Appearance-from-Motion Embedding}

\begin{figure}[t]
  \centering
    \begin{subfigure}[ht]{1\linewidth}
    \includegraphics[width=1\linewidth]{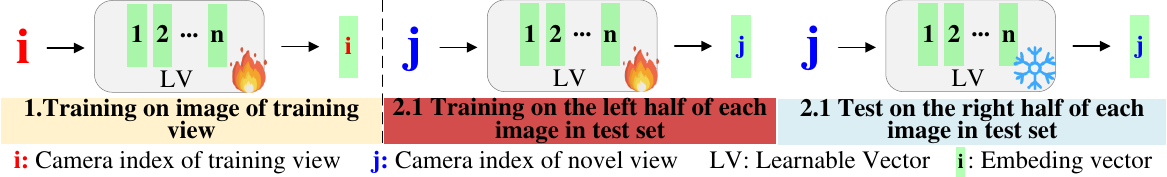}
    \caption{Appearance embedding in NeRF-W \cite{NeRF-W2021}. }
    \end{subfigure} 
    \begin{subfigure}[ht]{1\linewidth}
   \includegraphics[width=1\linewidth]{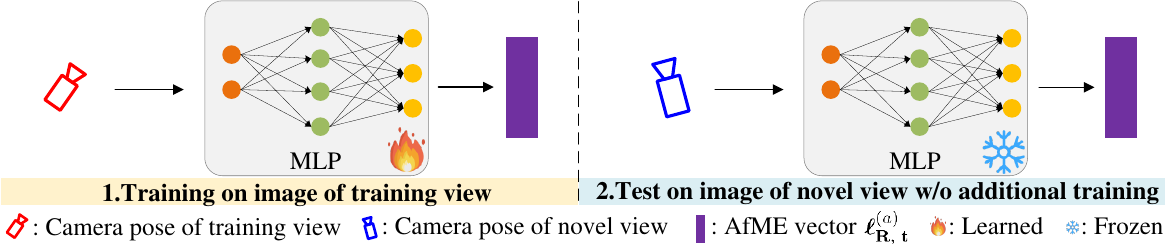}
   \caption{Our Appearance-from-Motion embedding. }
    \end{subfigure} 
   \caption{AE \cite{NeRF-W2021} and the proposed AfME. The differences between them are:  (1) AE uses image indexes as input, whereas AfME leverages camera poses.  (2) AE adopts learnable vectors (LV), while AfME utilizes a tiny MLP.  (3) Most critically, our AfME requires no additional training on novel views.  }
   \label{fig:afme}
\end{figure}
To further enhance the rendering quality of SEGS-SLAM, we observe a drop in the photorealistic mapping quality when suffering from appearance changes. AE is a proven solution in \cite{NeRF-W2021,Scaffold-GS2024} for handling such variations. However,  the limitation of the AE in \cite{NeRF-W2021} lies in involving each ground-truth (left half of the image) in the test set for training as shown in the mid of 
\cref{fig:afme} (a).
To address this issue, we propose Appearance-from-Motion embedding (AfME), which employs a lightweight Multilayer Perceptron (MLP) $M_{\theta_a}$ to learn a shared appearance representation. As illustrated in \cref{fig:afme} (b), the input of the encoder $M_{\theta_a}$ is the the camera pose $(\mathrm{\mathbf{R}},\ \mathrm{\mathbf{t}})$. The MLP encodes the pose and outputs an embedding vector $\bm{\ell}^{(a)}_{\mathrm{\mathbf{R}},\ \mathrm{\mathbf{t}}}$, as shown below:
\begin{equation}
    \bm{\ell}^{(a)}_{\mathrm{\mathbf{R}},\ \mathrm{\mathbf{t}}} = M_{\theta_a}(\mathrm{\mathbf{R}},\ \mathrm{\mathbf{t}}).
  \label{eq:embedding}    
\end{equation}
Subsequently, the embedding vector $\bm{\ell}^{(a)}_{\mathrm{\mathbf{R}},\ \mathrm{\mathbf{t}}}$ is fed into the color decoder $M_C$. After introducing AfME, the color prediction of 3D Gaussians changes from \cref{eq:color_mlp1} to:
\begin{equation}
\label{eq:color_mlp}  
\{c_0,\dots,c_{k-1}\} = M_C(\hat{f}_v,\ \delta_{vc},\ \vec{\mathrm{\mathbf{d}}}_{vc},\ \bm{\ell}^{(a)}_{\mathrm{\mathbf{R}},\ \mathrm{\mathbf{t}}}).
\end{equation}
More details are in \cref{sec:sup_afme} of supplementary material. We choose camera poses as inputs for several reasons:  
1)  Similar to image indices, camera poses are unique for each view.  
2) Camera poses naturally represent spatial information, enabling AfME to predict appearance from spatial context.  
3) Camera poses are more continuous than image indices. 

\begin{figure}[t]
  \centering

    \begin{subfigure}[ht]{1\linewidth}
    \includegraphics[width=1\linewidth]{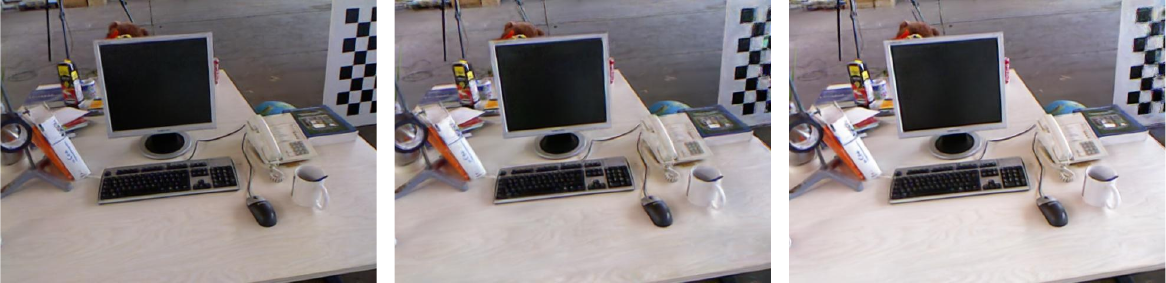}
   \caption{
   Same-view renderings with different appearance embedded by AfME.
   }
    \end{subfigure}   
    \begin{subfigure}[ht]{1\linewidth}
    \includegraphics[width=1\linewidth]{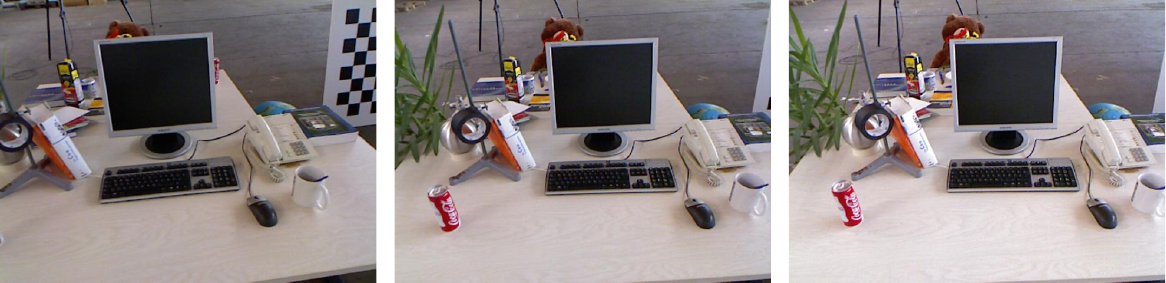}
   \caption{
   Renderings from different views correspond to different appearances.}
    \end{subfigure}   
   \caption{
   The visualization of AfME controlling appearance. The rendering viewpoints in the top three images above are same, and only the input to AfME has been changed. The input poses of the AfME in the top-row images correspond to those in the bottom-row images. The results show that only the color and illumination have changed, while the geometry is fixed. }
   \label{fig:apperance}
\end{figure}

We adopt AfME to encode the scene appearance into the continuous pose space. Through training on the training set, AfME learns the mapping between appearance and camera pose, enabling it to predict the appearance for novel views. We conduct an experiment to demonstrate this. After training, we fix $\hat{f}_v,\ \delta_{vc},\ \vec{\mathrm{\mathbf{d}}}_{vc}$ in the input of  \cref{eq:color_mlp} and vary only the pose $\{\mathrm{\mathbf{R}},\ \mathrm{\mathbf{t}}\}$ fed into AfME.  As shown in \cref{fig:apperance}, the lighting conditions under the same view change consistently with the pose $\{\mathrm{\mathbf{R}},\ \mathrm{\mathbf{t}}\}$, matching the illumination of corresponding viewpoints. This demonstrates that the illumination conditions can be embedded by AfME effectively.

\subsection{Frequency Pyramid Regularization }
A minor improvement is the frequency pyramid regularization (FPR), which leverages multi-scale frequency representation to enhance the reconstruction of high-frequency details in the scene. To achieve this, we apply bilinear interpolation to downsample both the render images $I_r$ and the ground truth images $I_g$. Let $s \in \mathcal{S} = \{s_0, s_1, \ldots, s_n\}$ denote the scale of an image.  We apply a 2D Fast Fourier Transform (FFT) to obtain the frequency spectra $\mathcal{F}(I_r^{s})(u, v), \mathcal{F}(I_g^{s})(u, v)$. The loss $\mathcal{L}_{hf}$ is computed as
\begin{align}
&\mathcal{L}_{hf} =\sum_{s \in \mathcal{S}} \frac{1}{\mathcal{N}} \lambda_s \sum_{u, v} \left| F_{hf, r}^{s}(u, v) - F_{hf, g}^{s}(u, v) \right| \label{eq:loss_fre}, \\
&F_{hf, i}^{s}(u, v) = H_{hf}(u, v) \cdot \mathcal{F}(I_i^{s})(u, v),\ i\in \{r, g\} ,
\end{align} \\
where $F_{hf, r}^{s}(u, v), F_{hf, g}^{s}(u, v)$ is the high-frequency extracted by a high-pass filter $H_{hf}(u, v)$,  $\mathcal{N} =HW$ denotes the image size, and $\lambda_s$ represents the weight of each scale. More details are in \cref{sec:sup_fpr} of supplementary material.


\subsection{Losses Design}

The optimization of the learnable parameters, the MLP $M_\alpha$, $M_c$, $M_q$, $M_s$, and $M_{\theta_a}$, are achieved by minimizing the L1 loss $\mathcal{L}_1$, SSIM term  \cite{SSIM} $\mathcal{L}_{\text{SSIM}}$, frequency regularization $\mathcal{L}_{hf}$, and volume regularization \cite{volumetric} $\mathcal{L}_{\text{vol}}$ between the rendered images and the ground truth images, denoted as
\begin{equation}
\mathcal{L} = (1-\lambda)\mathcal{L}_1 +  \lambda(1-\mathcal{L}_{\text{SSIM}}) +\lambda_{\text{vol}}\mathcal{L}_{\text{vol}} + \lambda_{hf} \mathcal{L}_{hf}.
\label{eq:loss}
\end{equation}
Following \cite{Scaffold-GS2024}, we also incorporate $\mathcal{L}_{\text{vol}}$.

\begin{table*}
\footnotesize
  \centering
   \setlength\tabcolsep{14pt} 
  \begin{tabular}{@{}l|ccc|ccc} 
    \hline   
   Datasets ({\bf Camera}) & \multicolumn{3}{c|}{Replica ({\bf RGB-D})} & \multicolumn{3}{c}{TUM RGB-D  ({\bf RGB-D})}  \\
    Method &PSNR $\uparrow$ &SSIM $\uparrow$ &LPIPS $\downarrow$ &PSNR $\uparrow$ &SSIM $\uparrow$ &LPIPS $\downarrow$  \\
    \hline       
       MonoGS \cite{MonoGS2024} & 36.81 & 0.964 & 0.069  & \cellcolor{lightorange}{24.11} &	\cellcolor{lightorange}{0.800}&	0.231  \\
       Photo-SLAM \cite{Photo-SLAM2024}  & 35.50 & 0.949 & 0.056 &	20.99 &	0.736  & \cellcolor{lightyellow}{0.213} 	 \\
       Photo-SLAM-30K & \cellcolor{lightyellow}{36.94}  & 0.952 & \cellcolor{lightorange}{0.040}  & 21.73 &0.757 &\cellcolor{lightorange}{0.186}\\
       RTG-SLAM \cite{RTG-SLAM2024} & 32.79 & 0.918 & 0.124  &	16.47 &	0.574 &	0.461\\
       GS-SLAM$^*$ \cite{GS-SLAM2024} & 34.27 & \cellcolor{lightred}{\bf 0.975} & 0.082  & - & - & - \\
       SplaTAM \cite{SplaTAM2024} & 33.85 & 0.936 & 0.099 & \cellcolor{lightyellow}{21.41} & \cellcolor{lightyellow}{0.764} & {0.265}  \\
       SGS-SLAM \cite{Sgs-slam2024} & 33.96 & \cellcolor{lightorange}{0.969} & 0.099 & - & - & -  \\
       GS-ICP SLAM \cite{GS-ICPSLAM2024} & \cellcolor{lightorange}{37.14} & \cellcolor{lightyellow}{0.968} & \cellcolor{lightyellow}{0.045}& 17.81 & 0.642 & 0.361 \\
       \hline
       {\bf Ours} &\cellcolor{lightred}{\bf 39.42} & \cellcolor{lightred}{\bf 0.975} & \cellcolor{lightred}{\bf 0.021}  &	\cellcolor{lightred}{\bf 26.03} &	\cellcolor{lightred}{\bf 0.843} & \cellcolor{lightred}{\bf 0.107}   \\
    \hline
  \end{tabular}
  \caption{Quantitative evaluation of our method compared to SOTA methods for {\bf RGB-D} camera on Replica and TUM RGB-D datasets. Best results are marked as  \colorbox{lightred}{\bf best score}, \colorbox{lightorange}{second best score} and \colorbox{lightyellow}{third best score}. GS-SLAM$^*$ denotes the result of GS-SLAM is taken from \cite{GS-SLAM2024}, all others are obtained in our experiments. '-' denotes the system does not provide valid results.}
  \label{tab:rendering_rgbd}
\end{table*}

\begin{figure*}[ht]
	\centering
    \begin{subfigure}[t]{1\linewidth}
        \includegraphics[width=1\linewidth]{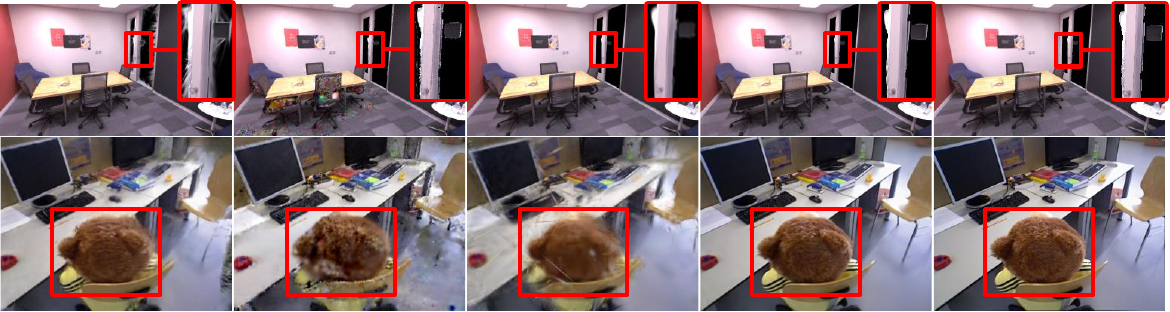}
    \end{subfigure}
    \begin{subfigure}[t]{0.19\linewidth}
        \vspace{-0.5cm}
        \caption{GS-ICP SLAM \cite{GS-ICPSLAM2024}}
    \end{subfigure}
    \begin{subfigure}[t]{0.19\linewidth}
        \vspace{-0.5cm}
        \caption{SplaTAM \cite{SplaTAM2024}}
    \end{subfigure}
    \begin{subfigure}[t]{0.19\linewidth}
        \vspace{-0.5cm}
        \caption{RTG-SLAM \cite{RTG-SLAM2024}}
    \end{subfigure}
    \begin{subfigure}[t]{0.19\linewidth}
        \vspace{-0.5cm}
        \caption{{\bf Ours}}
    \end{subfigure}
    \begin{subfigure}[t]{0.19\linewidth}
        \vspace{-0.5cm}
        \caption{Ground Truth}
    \end{subfigure}
	\caption{We show comparisons of ours to SOTA methods for \textbf{RGB-D} camera. The top scene is \textit{office2} from the Replica datasets, and the bottom is \textit{fr3/office} from TUM RGB-D datasets. Non-obvious differences in quality are highlighted by insets.}
	\label{fig:rendering_rgbd}
\end{figure*}

\begin{table*}[t]
\footnotesize
  \centering
   \setlength\tabcolsep{10pt} 
  \begin{tabular}{@{}l|ccc|ccc|ccc} 
    \hline   
   Datasets ({\bf Camera}) & \multicolumn{3}{c|}{Replica ({\bf Mono})} & \multicolumn{3}{c|}{TUM RGB-D ({\bf Mono})} & \multicolumn{3}{c}{EuRoC ({\bf Stereo})} \\
   method &  PSNR $\uparrow$ &SSIM $\uparrow$   &LPIPS $\downarrow$  
        &  PSNR $\uparrow$ &SSIM $\uparrow$   &LPIPS  $\downarrow$  
        &  PSNR $\uparrow$ &SSIM $\uparrow$   &LPIPS $\downarrow$  \\
    \hline  
       MonoGS \cite{MonoGS2024} & 28.34 & 0.878 & 0.256   &	21.00 &	0.705 & 0.393  & \cellcolor{lightorange}{22.60} & \cellcolor{lightorange}{0.789}  &	\cellcolor{lightorange}{0.274}  \\
       Photo-SLAM \cite{Photo-SLAM2024}  & 33.60 & 0.934 & 0.077   &	20.17 &	0.708 & 0.224  & 11.90  & 0.409 & 0.439 \\
       Photo-SLAM-30K & \cellcolor{lightorange}{36.08} & \cellcolor{lightorange}{0.947} & \cellcolor{lightorange}{0.054}  & \cellcolor{lightorange}{21.06} & \cellcolor{lightorange}{0.733} & \cellcolor{lightorange}{0.186} 	& 11.77 & 0.405 & 0.430\\
       \hline
       {\bf Ours} & \cellcolor{lightred}{\bf 37.96} &\cellcolor{lightred}{\bf 0.964} & \cellcolor{lightred}{\bf 0.037}  &\cellcolor{lightred}{\bf 25.17} &\cellcolor{lightred}{\bf 0.825}  & \cellcolor{lightred}{\bf 0.122} & \cellcolor{lightred}{\bf 23.64} & \cellcolor{lightred}{\bf0.791} & \cellcolor{lightred}{\bf 0.182} \\
   \hline
  \end{tabular}
  \caption{Quantitative evaluation of our method compared to SOTA methods for {\bf Monocular (Mono)} and {\bf Stereo} cameras on Replica, TUM RGB-D, and EuRoC MAV datasets. Best results are marked as  \colorbox{lightred}{\bf best score} and \colorbox{lightorange}{second best score}.}
  \label{tab:rendering_mono_stereo}
\end{table*}

\section{Experiment}

\subsection{Experiment Setup}



\noindent\textbf{Implementation.} Our SEGS-SLAM is fully implemented using the LibTorch framework with C++ and CUDA.  The training and rendering of 3D Gaussians involves three key modules: SEPM, AfME, and FPR, operating as a parallel thread alongside the localization and geometric mapping process. SEGS-SLAM trains the 3D Gaussians using only keyframe images, point clouds, and poses, where keyframes are selected based on co-visibility. In each iteration, SEGS-SLAM randomly samples a viewpoint from the current set of keyframes. We use the images and poses of keyframes as the training set, while the remaining images and poses serve as the test set. Moreover, following FreGS \cite{FreGS2024}, we activate FPR once the structure of anchor points stabilizes and terminate it based on the completion of anchor point densification. The scale level of FPR is set to 3. Except for the non-open-source GS-SLAM \cite{GS-SLAM2024}, all methods compared in this paper are run on the same machine using their official code. The machine is equipped with an NVIDIA RTX 4090 GPU and a Ryzen 5995WX CPU. By default, our method runs for 30K iterations.  The voxel size $\epsilon$ is $ 0.001 \, \text{m}$. For~\cref{eq:loss}, we set $\lambda=0.2, \lambda_{\text{vol}}=0.01, \lambda_{hf}=0.01$.

\noindent\textbf{Baselines.} We first list the baseline methods used to evaluate photorealistic mapping. For monocular and stereo cameras, we compare our method with Photo-SLAM \cite{Photo-SLAM2024}, MonoGS \cite{MonoGS2024}, and Photo-SLAM-30K. For RGB-D cameras, we additionally include comparisons with RTG-SLAM \cite{RTG-SLAM2024}, GS-SLAM \cite{GS-SLAM2024}, SplaTAM \cite{SplaTAM2024}, SGS-SLAM \cite{Sgs-slam2024}, and GS-ICP SLAM \cite{GS-ICPSLAM2024}, all of which represent SOTA SLAM methods based on 3D-GS. To ensure fairness, we set the maximum iteration limit to 30K for all methods, following the original 3D-GS \cite{3DGS2023}. Photo-SLAM-30K refers to Photo-SLAM \cite{Photo-SLAM2024} with a fixed iteration count of 30K.  For camera pose estimation, we also compare ours with ORB-SLAM3 \cite{ORB-SLAM32021} and DROID-SLAM \cite{DROID-SLAM2021}. 

\noindent\textbf{Metrics.} We follow the evaluation protocol of MonoGS \cite{MonoGS2024} to assess both camera pose estimation and novel view synthesis. For camera pose estimation, we report the root mean square error (RMSE) of the absolute trajectory error (ATE) \cite{grupp2017evo} for all frames. For photorealistic mapping, we report standard rendering quality metrics, including PSNR, SSIM, and LPIPS \cite{lpips}. To evaluate the photorealistic mapping quality, we only calculate the average metrics over novel views for all methods. We report the average across five runs for all methods. \textit{To ensure fairness, no training views are included in the evaluation, and for all RGB-D SLAM methods, no masks are applied to either the rendered or ground truth images during metric calculation.} As a result, the reported metrics for Photo-SLAM \cite{Photo-SLAM2024} are slightly lower than those in the original paper, as they averages both novel and training views. Similarly, the metrics of SplaTAM  \cite{SplaTAM2024}, SGS-SLAM \cite{Sgs-slam2024}, and GS-ICP SLAM \cite{GS-ICPSLAM2024} are slightly lower than reported, as the original methods use a mask to exclude outliers for both the rendered and ground truth images based on anomalies in the depth image.

\noindent\textbf{Datasets.} Following \cite{Photo-SLAM2024, GS-SLAM2024, GS-ICPSLAM2024, SplaTAM2024, MonoGS2024, CG-SLAM2024, RTG-SLAM2024, Point-SLAM2023, iMAP2021, Sgs-slam2024}, we evaluate all methods on all sequences of the Replica dataset \cite{Replica2019} for monocular and RGB-D cameras. Following \cite{Photo-SLAM2024, GS-SLAM2024, GS-ICPSLAM2024,  MonoGS2024, RTG-SLAM2024, iMAP2021, NICE-SLAM2022}, we use the \textit{ fr1/desk}, \textit{fr2/xyz}, and \textit{fr3/office} sequences of the TUM RGB-D dataset \cite{TUMRGB-D2012} for monocular and RGB-D cameras. Following \cite{Photo-SLAM2024}, we use the \textit{MH01}, \textit{MH02}, \textit{V101}, and \textit{V201} sequences of the EuRoC MAV dataset \cite{EuRoC2016} for stereo cameras.

\begin{table}
\scriptsize
\fontsize{6.5}{10}\selectfont
  \centering
   \setlength\tabcolsep{1pt} 
  \begin{tabular}{@{}l|ccc|ccc|c} 
    \hline   
     Camera Type & \multicolumn{3}{c|}{ RGB-D} & \multicolumn{3}{c|}{Monocular} & {Stereo} \\
    Datasets & Replica & TUM R  & Avg. & Replica & TUM R & Avg. & EuRoC \\
   Method &  RMSE $\downarrow$ &RMSE $\downarrow$   &RMSE $\downarrow$ &  RMSE $\downarrow$ &RMSE $\downarrow$ &  RMSE $\downarrow$ &RMSE $\downarrow$  \\
    \hline  
        {ORB-SLAM3 \cite{ORB-SLAM32021}}  & 1.780  &2.196	&  1.988   &51.744 &46.004  & 48.874 &\cellcolor{lightyellow}{10.907}	 \\
        {DRIOD-SLAM \cite{DROID-SLAM2021}} &74.264  &74.216  		& 74.24 &76.600 & \cellcolor{lightyellow}{1.689}    & \cellcolor{lightyellow}{39.145} & \cellcolor{lightred}{\bf 1.926} 	 \\
        {MonoGS \cite{MonoGS2024}}& 0.565 & \cellcolor{lightorange}{1.502} &  \cellcolor{lightyellow}{1.033} & \cellcolor{lightyellow}{37.054}  & 4.009 &  63.437 &	49.241 \\
        {Photo-SLAM \cite{Photo-SLAM2024}} &0.582 &1.870	&  1.226	& \cellcolor{lightorange}{ 0.930} & \cellcolor{lightorange}{ 1.539} &  \cellcolor{lightorange}{ 1.235} &	11.023\\
        RTG-SLAM \cite{RTG-SLAM2024}& \cellcolor{lightorange}{0.191} & \cellcolor{lightred}{\bf0.985}  &   \cellcolor{lightred}{\bf 0.581} & - 	  &- & - 	&- \\
        GS-SLAM$^*$ \cite{GS-SLAM2024} &0.500 &3.700	 & 2.100  & - 	  &- &  -	&-\\
        {SplaTAM \cite{SplaTAM2024}}& \cellcolor{lightyellow}{0.343} & 4.215 & 2.279  & - 	  &- &  -	&- \\
        SGS-SLAM \cite{Sgs-slam2024} & 0.365 & - & -  & - 	  &- &  -	&- \\
        {GS-ICP SLAM \cite{GS-ICPSLAM2024}}& \cellcolor{lightred}{\bf0.177} &2.921	&1.549	& -  &- &-	&-\\
        \hline
        {\bf Ours} & 0.430 &  \cellcolor{lightyellow}{1.528} & \cellcolor{lightorange}{0.979}  & \cellcolor{lightred}{\bf 0.833} &\cellcolor{lightred}{\bf 1.505} &  \cellcolor{lightred}{\bf 1.169} & \cellcolor{lightorange}{ 7.462} \\
    \hline
  \end{tabular}
  \caption{Camera tracking result on Replica, TUM RGB-D (TUM R), and EuRoC MAV datasets for Monocular, stereo, and RGB-D cameras.  {\bf RMSE of ATE (\rm{cm})} is reported. The best results are marked as  \colorbox{lightred}{\bf best score}, \colorbox{lightorange}{second best score} and \colorbox{lightyellow}{third best score}.  '-' denotes that the system does not provide valid results.}
  \label{tab:tracking}
\end{table}

\begin{figure*}[ht]
	\centering
    \begin{subfigure}[ht]{1\linewidth}
        \includegraphics[width=1\linewidth]{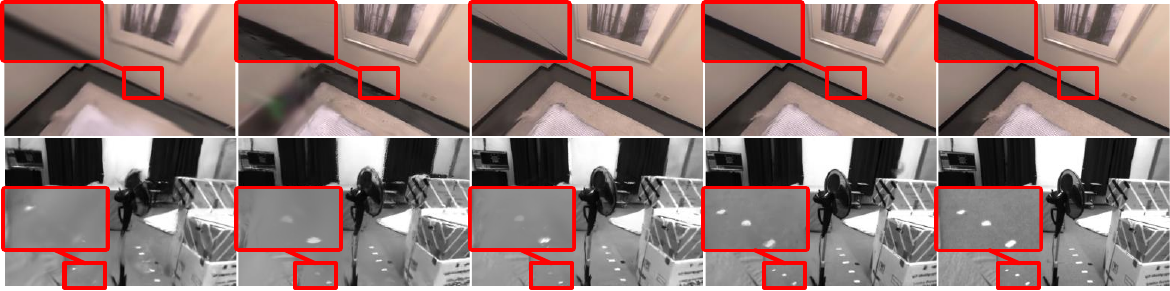}
    \end{subfigure}
    \begin{subfigure}[ht]{0.19\linewidth}
           \vspace{-0.15cm}
        \caption{MonoGS \cite{MonoGS2024}}
    \end{subfigure}
    \begin{subfigure}[ht]{0.19\linewidth}
            \vspace{-0.15cm}
        \caption{Photo-SLAM \cite{Photo-SLAM2024}}
    \end{subfigure}
    \begin{subfigure}[ht]{0.19\linewidth}
            \vspace{-0.15cm}
        \caption{Photo-SLAM-30k}
    \end{subfigure}
    \begin{subfigure}[ht]{0.19\linewidth}
            \vspace{-0.15cm}
        \caption{{\bf Ours}}
    \end{subfigure}
    \begin{subfigure}[ht]{0.19\linewidth}
           \vspace{-0.15cm}
        \caption{Ground Truth}
    \end{subfigure}
	\caption{We show comparisons of ours to SOTA methods for \textbf{Monocular} and \textbf{Stereo}  cameras. The top scene is \textit{room1} from the Replica dataset, and the bottom is \textit{V201} from the EuRoC MAV dataset. Non-obvious differences in quality are highlighted by insets.}
	\label{fig:rendering_mono_stereo}
\end{figure*}

\subsection{Results Analysis}
\textbf{Camera Tracking Accuracy.}
As shown in \cref{tab:tracking}, our method demonstrates competitive accuracy in tracking for monocular, stereo, and RGB-D cameras when compared with SOTA methods. This highlights the advantage of indirect visual SLAM in terms of localization accuracy.

\noindent\textbf{Novel View synthesis.}
The quantitative rendering results for novel views in \textit{RGB-D} scenarios are shown in \cref{tab:rendering_rgbd}, where SEGS-SLAM significantly outperforms comparison methods, achieving the highest average rendering quality on both TUM RGB-D and Replica datasets. In the top of \cref{fig:rendering_rgbd}, it is evident that for the Replica dataset, only our method can accurately recover the contours of edge regions. The TUM RGB-D dataset presents a greater challenge compared with the Replica dataset, with highly cluttered scene structures and substantial lighting variations. GS-ICP SLAM \cite{GS-ICPSLAM2024}, a leading RGB-D SLAM method based on 3D-GS, achieves the second-highest rendering accuracy on the Replica dataset. However, it relies heavily on depth images. As shown in the bottom of \cref{fig:rendering_rgbd}, GS-ICP SLAM \cite{GS-ICPSLAM2024} and other methods perform poorly on the TUM RGB-D dataset. Our SEGS-SLAM better reconstructs scene structure and lighting variations, benefiting from our SEPM and the AfME.

\cref{tab:rendering_mono_stereo} presents quantitative rendering results for \textit{monocular} scenarios, where SEGS-SLAM surpasses other methods. Notably, SEGS-SLAM continues to significantly outperform comparison methods on the TUM RGB-D dataset. Importantly, compared with RGB-D scenarios, MonoGS \cite{MonoGS2024} experiences a sharp decline. The top of \cref{fig:rendering_mono_stereo} further demonstrates that on the Replica dataset, our method effectively models high-frequency details more realistically in regions such as the edge of the wall.

Moreover, our method remains effective in \textit{stereo} scenarios. The corresponding quantitative results for realistic mapping are recorded in \cref{tab:rendering_mono_stereo}, where our method achieves the highest rendering quality, surpassing the current SOTA method, MonoGS \cite{MonoGS2024}. As shown in the bottom of \cref{fig:rendering_mono_stereo}, our approach better reconstructs the global structure and local details of the scene. We highlight that a key factor enabling our method to achieve superior performance across different camera types and datasets is the proposed SEPM. Its enhancement to rendering quality is highly generalizable, as further demonstrated in ablation studies.

\subsection{Ablation Studies}

\begin{figure}[t]
	\centering
    \begin{subfigure}[ht]{1\linewidth}
    \includegraphics[width=1\linewidth]{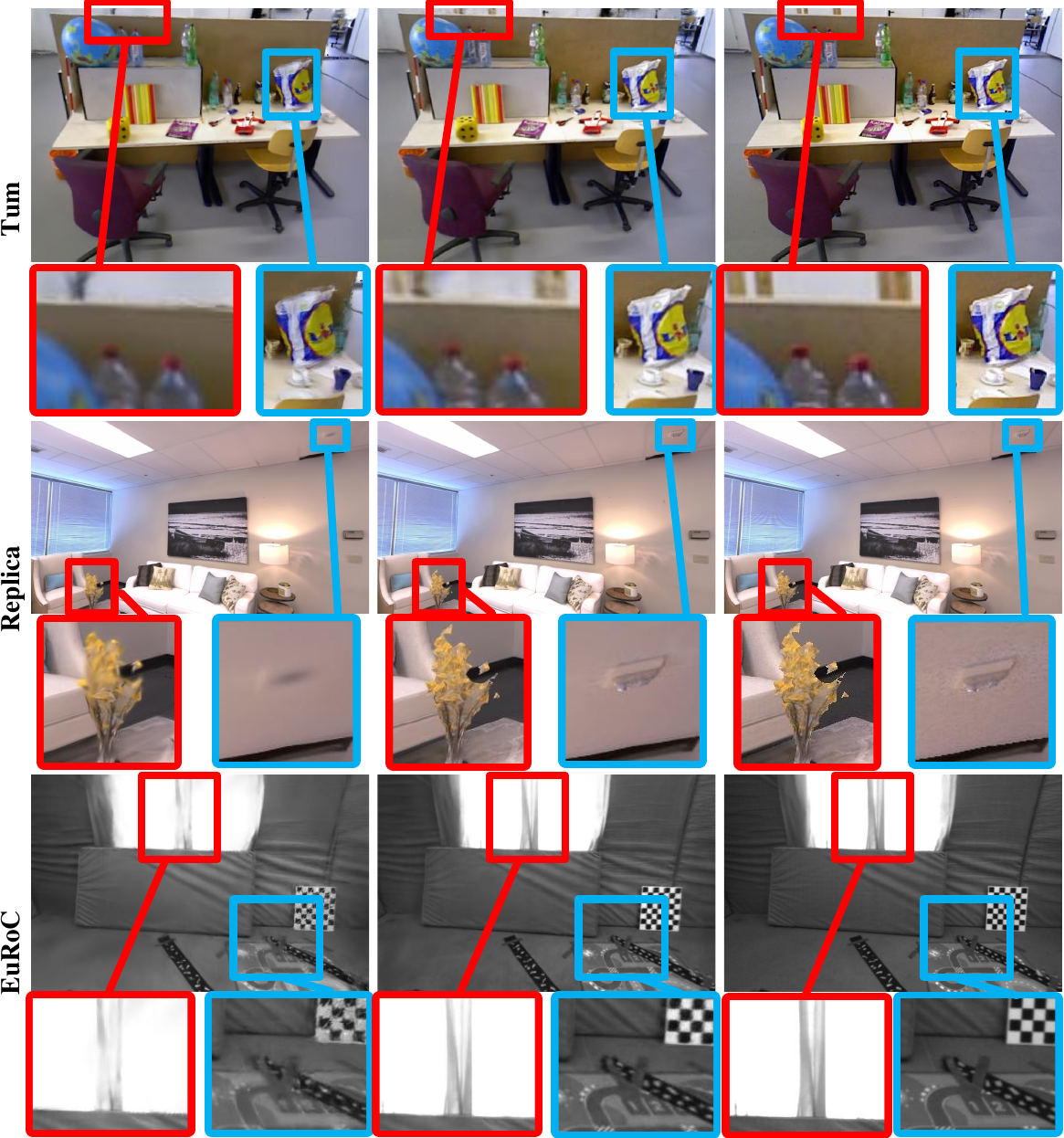}
    \end{subfigure}
    \begin{subfigure}[ht]{0.32\linewidth}
            \vspace{-0.15cm}
        \caption*{ Ours w/o FPR, AfME, SEPM}
    \end{subfigure}   
    \begin{subfigure}[ht]{0.32\linewidth}
            \vspace{-0.15cm}
        \caption*{Ours w/o FPR, AfME}
    \end{subfigure}    
    \begin{subfigure}[ht]{0.32\linewidth}
            \vspace{-0.15cm}
        \caption*{  Ground Truth}
    \end{subfigure}
	\caption{Ablation of SEPM. The figure presents qualitative results on the TUM, Replica, and EuRoC datasets. It is evident that, across all three datasets, incorporating SEPM enables the reconstruction of more complete and accurate structure.  
   }
	\label{fig:abl_structure}
\end{figure}

\noindent\textbf{Structure-Enhanced photorealistic mapping.}
To evaluate the impact of SEPM on photorealistic mapping metrics, we additionally train two variants of our method: one without SEPM, AfME, and FPR, and another without AfME and FPR. The variant without SEPM, AfME, and FPR directly uses the original 3D-GS \cite{3DGS2023}. As shown in \cref{tab:ablation1}, rows (1) and (2), introducing SEPM consistently yields significant improvements in rendering quality across all scenes. This validates the effectiveness of enhancing the exploitation of latent structure. Moreover, it demonstrates that SEPM has strong generalization across diverse camera inputs and datasets, inspiring future research. Notably, with only SEPM, our rendering metrics surpass existing SOTA methods. To further illustrate the benefits of SEPM, \cref{fig:abl_structure} presents visual comparisons, where a more complete and accurate scene structure leads to superior rendering quality.

\begin{figure}[t]
	\centering
    \begin{subfigure}[ht]{1\linewidth}
    \includegraphics[width=1\linewidth]{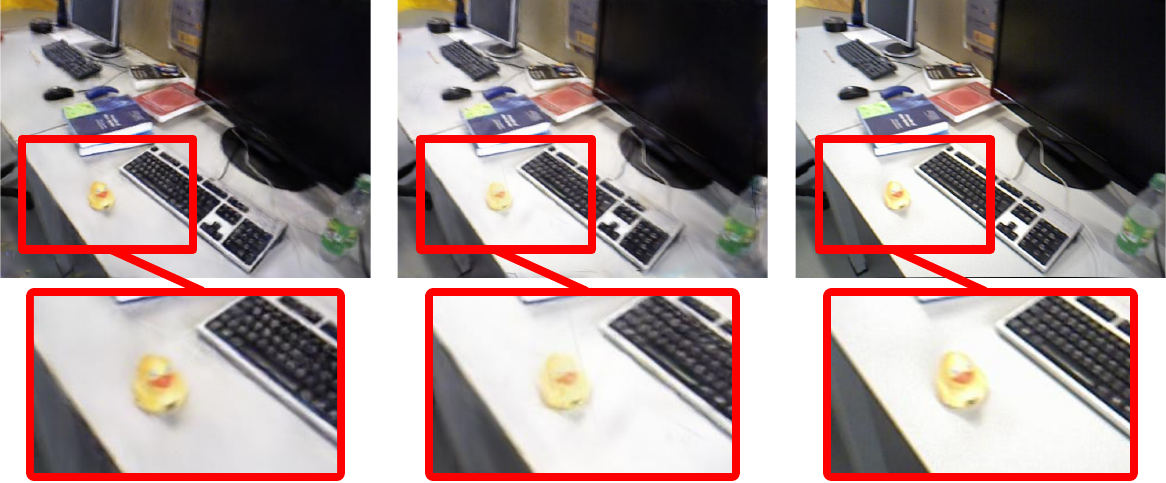}
    \end{subfigure}   

    \begin{subfigure}[ht]{0.32\linewidth}
            \vspace{-0.15cm}
        \caption*{Ours w/o AfME}
    \end{subfigure}   
    \begin{subfigure}[ht]{0.32\linewidth}
            \vspace{-0.15cm}
        \caption*{{\bf Ours}}
    \end{subfigure}    
    \begin{subfigure}[ht]{0.32\linewidth}
            \vspace{-0.15cm}
        \caption*{ Ground Truth}
    \end{subfigure}  
    \begin{subfigure}[ht]{1\linewidth}
    \includegraphics[width=1\linewidth]{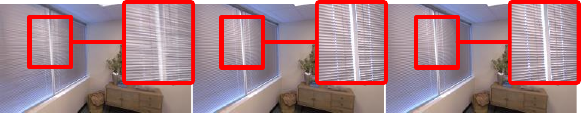}
    \end{subfigure} 
    \begin{subfigure}[ht]{0.32\linewidth}
            \vspace{-0.15cm}
        \caption*{ Ours w/o FPR}
    \end{subfigure}   
    \begin{subfigure}[ht]{0.32\linewidth}
            \vspace{-0.15cm}
        \caption*{{\bf Ours}}
    \end{subfigure}    
    \begin{subfigure}[ht]{0.32\linewidth}
            \vspace{-0.15cm}
        \caption*{  Ground Truth}
    \end{subfigure}  
	\caption{Ablation of AfME (\textit{Top}) and FPR (\textit{Bottom}). It is evident that with the introduction of AfME, the lighting conditions at novel views are accurately predicted, indicating the effectiveness of our AfME. The result without AfME is darker than the ground truth.}
	\label{fig:ablation_fpr}
\end{figure}

\noindent\textbf{Appearance-from-Motion embedding.}
To evaluate the impact of the proposed AfME, we train an additional model for our method without AfME. In the rows (4) and (5) of \cref{tab:ablation1}, our full method (5) outperforms the model without AfME (4) in terms of PSNR scores. As shown in the top of \cref{fig:ablation_fpr}, the AfME effectively predicts the lighting conditions of novel views. On the Replica dataset, the improvements from AfME are relatively modest. Replica is an easier dataset, in which PSNR already exceeds 37 without AfME, indicating that scene is well-reconstructed. Although incorporating AfME still yields a PSNR gain, the benefits are much more pronounced on the challenging TUM dataset, as clearly observed in both \cref{fig:ablation_fpr} and \cref{tab:ablation1}.



\noindent\textbf{Frequency pyramid regularization.} 
To evaluate the effect of the proposed FPR on photorealistic mapping metrics, we train an additional model for our method without FPR. As shown in \cref{tab:ablation1}, rows (3) and (5), our full method (5) surpasses the model without FPR (3) in terms of PSNR scores. Additionally, after applying the FPR, the model renders finer details in highly textured regions, as demonstrated by the curtain at the bottom of \cref{fig:ablation_fpr}.  

\begin{table}
\scriptsize
  \centering
   \setlength\tabcolsep{3.5pt} 
  \begin{tabular}{@{}ll|cc|cc|c} 
    \hline  
  \multicolumn{2}{l|}{ Camera type} & \multicolumn{2}{c|}{RGB-D} & \multicolumn{2}{c|}{Mono} & Stereo\\
   \multicolumn{2}{l|}{Datasets} & Replica & TUM R & Replica & TUM R & EuRoC \\
    \#& Method &  PSNR $\uparrow$ &  PSNR $\uparrow$  &  PSNR $\uparrow$ &  PSNR $\uparrow$  &  PSNR $\uparrow$  \\
    \hline   
    (1)& w/o FPR, AfME, SEPM & 36.07 & 21.73 & 35.46 & 21.10 &	11.76 \\
    (2)&w/o FPR, AfME & 38.98 & 24.20 & 36.31 & 23.54 &	22.91 \\
     (3)&  w/o FPR  & 39.18 & 25.04 & 36.44 & 24.91 &	23.52 \\
     (4)&  w/o AfME & 39.12 & 24.66 & 37.48 & 23.69 &	22.99 \\

     (5)&  {\bf Ours}  &  \cellcolor{lightred}{\bf 39.42} & \cellcolor{lightred}{\bf 26.03} &  \cellcolor{lightred}{\bf 37.96} & \cellcolor{lightred}{\bf 25.17} &	\cellcolor{lightred}{\bf 23.64} \\
    \hline
  \end{tabular}
  \caption{ Ablation Study on the {\bf key components} (1) - (5). The best results are marked as  \colorbox{lightred}{\bf best score}.}
  \label{tab:ablation1}
\end{table}

\subsection{Limitations}
One limitation of our method is that a poorly structured point cloud leads to a decline in photorealistic mapping quality. Additionally, while our method achieves real-time tracking and rendering at 17 and 400 FPS, respectively, it exhibits reduced rendering speed due to the increased number of 3D Gaussians used to model high-frequency details. Currently, AFME is only capable of handling static scenes. By encoding more complex inputs, it can tackle more sophisticated dynamic scenarios.

\section{Conclusion}
We propose a novel SLAM system with progressively refined 3D-GS, termed SEGS-SLAM. Experimental results show that our method surpasses SOTA methods in rendering quality across monocular, stereo, and RGB-D datasets. We demonstrate that by enhancing the utilization of the underlying scene structure, SEPM improves the visual quality of rendering. Furthermore, our proposed AfME and FPR effectively predict the appearance of novel views and refine the scene details, respectively. 

\noindent\textbf{Acknowledgements:} This work is supported  by the National Natural Science Foundation
of China under Grant 62233011.
{
    \small
    \bibliographystyle{ieeenat_fullname}
    \bibliography{main}
}
\clearpage
\setcounter{page}{1}
\maketitlesupplementary

\section{Overview}
\label{sec:sup_Overview}

The supplementary material is organized as follows: (1) \cref{sec:sup_fpr} introduces more details of FPR. (2) \cref{sec:sup_abl} presents additional  ablation studies. (3) \cref{sec:sup_realtim} provides real-time performance for all methods.  (4) \cref{sec:sup_Imple} provides additional implementation details, including the detailed pipeline for localization and geometry mapping (\cref{sec:sup_SLAM}), the MLP architecture used for structured 3D Gaussians (\cref{sec:sup_s3d}), the MLP structure for Appearance-from-Motion Embedding (\cref{sec:sup_afme}), and anchor point refinement (\cref{sec:sup_fpr}). (6) \cref{sec:sup_result} presents quantitative results for each scene and includes more comparative renderings.

\section{Details of FPR}
\label{sec:sup_fpr}
In scenes with simple structures, our structured 3D Gaussians can effectively model both structure and appearance changes. However, we observe that structured 3D Gaussians perform poorly in rendering high-frequency details, such as object edges and areas with complex textures. Hence, we propose the frequency pyramid regularization (FPR) technique, which effectively leverages multi-scale frequency spectra. Here, we introduce the frequency pyramid to improve the consistency of rendering details for the same object across varying viewpoint distances. Unlike FreGS \cite{FreGS2024}, we leverage only high-frequency information, as low-frequency components typically represent scene structure, which is already effectively captured by our structured 3D Gaussians as shown in \cref{tab:sup_ablation2}.

\begin{figure}[t]
	\centering
    \begin{subfigure}[ht]{1\linewidth}
    \includegraphics[width=1\linewidth]{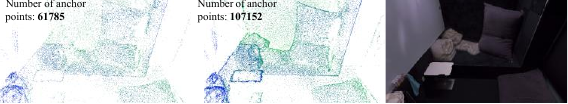}
    \end{subfigure}   
    \begin{subfigure}[ht]{0.32\linewidth}
            \vspace{-0.15cm}
        \caption{Ours w/o FPR}
    \end{subfigure}   
    \begin{subfigure}[ht]{0.32\linewidth}
            \vspace{-0.15cm}
        \caption{{\bf Ours}}
    \end{subfigure}    
    \begin{subfigure}[ht]{0.32\linewidth}
            \vspace{-0.15cm}
        \caption{Ground Truth}
    \end{subfigure}   
	\caption{Visualization of anchor points after 30K iterations. Increasing the number of Gaussians along edges improves the rendering quality. In the figure, 3D Gaussians are $k=10$ times the number of anchor points. }
	\label{fig:anchor_point}
\end{figure}

The primary effect of FPR is to guide the densification of anchor points. Specifically, when the average gradient of all Gaussians within a voxel exceeds a threshold $ \tau_g$, a new anchor point is added at the center of the voxel. Consequently, if a high-frequency region in the scene exhibits a substantial discrepancy between output rendering and ground truth, the total loss includes the frequency regularization $\mathcal{L}_{hf}$ increases, pushing the average gradient beyond $ \tau_g$. Then, more new anchors are added in the region, and scene edges become sharper, as shown in \cref{fig:anchor_point}. Thus, the high-frequency details in the scene are refined by FPR.

\section{Additional Ablation Studies}
\label{sec:sup_abl}
\noindent\textbf{Scale level of FPR.} We think that multiple scales improve the consistency under varying observation distances. The scale level of FPR is set to 3. Results are shown in  \cref{tab:sup_ablation1}.

\begin{table}
\scriptsize
  \centering
  \begin{tabular}{@{}l|cc|cc|c} 
       \hline
  { Camera type} & \multicolumn{2}{c|}{RGB-D} & \multicolumn{2}{c|}{Mono} & Stereo\\
  {Datasets} & Replica & TUM R & Replica & TUM R & EuRoC \\
     Scale level   &  PSNR  $\uparrow$ &  PSNR $\uparrow$  &  PSNR $\uparrow$  &  PSNR $\uparrow$   &  PSNR $\uparrow$ \\
       \hline
   1 & 39.14 & 25.82 & 37.65 & 23.83 & 23.58 \\
    2 & 39.29 & 25.53 & 37.55 & 23.77 & 23.57 \\
    4 & 39.34 & 26.02 & 37.51 & 24.75 & 23.42 \\
 3 (\textbf{Ours} )  & \textbf{39.42} & \textbf{26.03} & \textbf{37.96} & \textbf{25.17} & \textbf{23.64}\\
       \hline
  \end{tabular}
  \caption{ Ablation Study on the scale level of FPR.}
  \label{tab:sup_ablation1}
\end{table}

\noindent\textbf{The low-frequency component in FPR.} In our experiments, we find that the low-frequency component of FPR conflicts with structured 3D Gaussians, resulting in degradation. The best result is using only the high-frequent component in FPR, as shown in  \cref{tab:sup_ablation2}.

\begin{table}
  \centering
   \scriptsize
  \begin{tabular}{@{}l|cc|cc|c} 
    \hline  
  { Camera type} & \multicolumn{2}{c|}{RGB-D} & \multicolumn{2}{c|}{Mono} & Stereo\\
  {Datasets} & Replica & TUM R & Replica & TUM R & EuRoC \\
     Metric &  PSNR  $\uparrow$  &  PSNR $\uparrow$ &  PSNR $\uparrow$  &  PSNR $\uparrow$  &  PSNR $\uparrow$\\ 
       \hline
   low freq. & 38.85 & 25.76 & 36.79 & 24.79 & 23.37\\
    low  \& high freq. & 39.23 & 25.84 & 37.72 & 24.84 & 23.27\\
 high freq. (\textbf{Ours}) & \textbf{39.42} & \textbf{26.03} & \textbf{37.96} & \textbf{25.17} & \textbf{23.64}\\
       \hline
  \end{tabular}
  \caption{ Ablation study on the low frequency component of FPR.}
  \label{tab:sup_ablation2}
\end{table}

\noindent\textbf{Replacement of key components.}
We train two additional models: one replaces AfME with the appearance embeddings (AE) from \textit{Scaffold}-GS \cite{Scaffold-GS2024}, and another replaces FPR with the single-scale frequency regularization (SFR). Our AfME differs from AE in two primary ways: 1) AfME uses camera poses as input, whereas AE uses camera indices; 2) AfME employs an MLP network structure, while AE utilizes an embedding layer. SFR refers to using only the original-scale image frequencies in \cref{eq:loss_fre}. The distinction between FPR and SFR lies in the use of multi-scale image frequencies in FPR.  As shown in \cref{tab:ablation1}, rows (1), (2), and (3), our full method (3) achieves the highest PSNR scores. This demonstrates that, compared with AE, our AfME is more effective in predicting appearance variations across a wide range of novel views, thus avoiding additional training on the test set. On the other hand, it also highlights that by introducing the frequency pyramid, the model maintain consistency in scene details across varying viewpoint distances, leading to superior rendering quality.

\begin{table}
\scriptsize
  \centering
   \setlength\tabcolsep{3.5pt} 
  \begin{tabular}{@{}ll|cc|cc|c} 
    \hline  
  \multicolumn{2}{l|}{ Camera type} & \multicolumn{2}{c|}{RGB-D} & \multicolumn{2}{c|}{Mono} & Stereo\\
   \multicolumn{2}{l|}{Datasets} & Replica & TUM R & Replica & TUM R & EuRoC \\
    \#& Method &  PSNR $\uparrow$ &  PSNR $\uparrow$  &  PSNR $\uparrow$ &  PSNR $\uparrow$  &  PSNR $\uparrow$  \\
    \hline
     (1) & replace AfME with AE  & 38.22 & 19.56 & 36.09 & 19.33 &	18.39 \\
     (2) &  replace FPR with SFR &  39.14 & 25.82 & 37.65 & 23.83 &	23.58  \\
     (3) &  {\bf Ours} &{\bf 39.42} & {\bf 26.03} & {\bf 37.96} & {\bf 25.17} &	 {\bf 23.64} \\
       \hline
  \end{tabular}
  \caption{ Ablation Study on the {\bf replacement of key components}. }
  \label{tab:sup_ablation3}
\end{table}

\section{Real-time performance} 
\label{sec:sup_realtim}
Our method, following Photo-SLAM, employs two parallel threads: \textit{Localization \& Geometry Mapping}  and \textit{3D-GS Mapping}. We note that only tracking and rendering are real-time. The runtime of all methods is provided in  \cref{tab:sup_realtime}.

\begin{table}
\tiny
  \centering
   \setlength\tabcolsep{1pt} 
\renewcommand{\arraystretch}{1}
  \begin{tabular}{@{}l|c|c|c|c|c|c|c} 
    \hline  
  Metric (RGB-D) & MonoGS & Photo-SLAM (-30K)  & RTG-SLAM  & SplaTAM & SGS-SLAM & GS-ICP SLAM  & {\bf Ours}  \\
       \hline
  Rendering FPS $\uparrow$  & 706 & {\bf 1562} (1439) & 447 & 531 & 486&	630&	400\\
   Tracking FPS $\uparrow$ & 1.33  & 30.30 ({\bf 30.87}) & 17.24 & 0.15 &	0.14&	30.32&	17.18 \\
   Mapping Time $\downarrow$ & 37m40s & {\bf 1m20s} {(6m32s)} & 12m03s & 3h45m & 4h05m&	{1m32s}&	11m14s\\
       \hline
  \end{tabular}
  \caption{ Real-time performance.}
  \label{tab:sup_realtime}
\end{table}

\section{Implementation details}
\label{sec:sup_Imple}

\subsection{System Overview}
Our system comprises two main modules: \textit{localization and geometry mapping} and \textit{progressively refined 3D Gaussian splatting (3D-GS)}. In our implementation, these two modules run in separate threads. The localization and geometry mapping module focuses on camera pose estimation and scene point cloud mapping. The progressively refined 3D-GS module takes the estimated keyframe poses and point clouds from the localization and geometry mapping module. Then the module incrementally completes the photorealistic mapping of the scene.

\subsection{Localization and Geometry Mapping}
\label{sec:sup_SLAM}
In our implementation, the localization and geometric mapping module consists of three main threads: \textit{tracking}, \textit{local mapping}, and \textit{loop closing}, along with an on-demand thread for\textit{ global bundle adjustment (BA)}. Specifically, the tracking thread performs a motion-only BA to optimize camera poses. The local mapping thread optimizes keyframe poses and map point clouds within a local sliding window via local BA. Lastly, the loop closing thread continuously checks for loop closures. If a loop is detected, a global BA is triggered to jointly optimize the camera poses of all keyframes and all points of the scene.

\noindent\textbf{Motion-only BA.} We optimize the camera orientation $\mathrm{\mathbf{R}}\in \mathrm{SO(3)}$ and position $t\in \mathbb{R}^3$ through motion-only BA. The camera poses $(\mathrm{\mathbf{R}}_\iota, \mathrm{\mathbf{t}}_\iota)$  are optimized by minimizing the reprojection error between the matched 3D points $\mathrm{\mathbf{P}}_\iota\in \mathbb{R}^3$ and 2D feature points $\mathrm{\mathbf{p}}_\iota$ within a sliding window:
\begin{align}
\{\mathrm{\mathbf{R}}_\iota,\mathrm{\mathbf{t}}_\iota\}&=  \sum_{\iota\in\mathcal{X}} \mathop{\mathrm{argmin}}\limits_{\mathrm{\mathbf{R}}_\iota,\mathrm{\mathbf{t}}_\iota}  \rho(\|p_\iota - \pi(\mathrm{\mathbf{R}_\iota}\mathrm{\mathbf{P}}_\iota +\mathrm{\mathbf{t}}_\iota) \|^2_{\Sigma_g}) 
\end{align}
where $\mathcal{X}$ represent the set of all matches, $\Sigma_g$ denote the covariance matrix associated with the keypoint's scale, $\pi$ is the projection function, and $\rho$ is the robust Huber cost function. 

\noindent\textbf{Local BA.} We perform a local BA by optimizing a set of covisible keyframes $\mathcal{K}_L$ alone with the set of points ${P}_L$ observed in those keyframes  as follows:
\begin{align}
\{\mathrm{\mathbf{P}}_m,\mathrm{\mathbf{R}}_l,\mathrm{\mathbf{t}}_l\} &= \mathop{\mathrm{argmin}}\limits_{\mathrm{\mathbf{P}}_m,\mathrm{\mathbf{R}}_l,\mathrm{\mathbf{t}}_l} \sum_{\kappa\in \mathcal{K}_L\cup \mathcal{K}_F}  \sum_{j\in \mathcal{X}_k} \rho(E(\kappa,j)) \\
E(\kappa, j) &= \|\mathrm{\mathbf{p}}_j - \pi(\mathrm{\mathbf{R}}_\kappa\mathrm{\mathbf{P}}_j+\mathrm{\mathbf{t}}_\kappa) \|^2_{\Sigma_g}
\end{align}
where  $m\in {P}_L$, $l\in \mathcal{K}_L$, $\mathcal{K}_F$ are all other keyframes, $\mathcal{X}_k$ is the set of matches between keypoints in a keyframe $\kappa$ and points in ${P}_L$. 

\noindent\textbf{Global BA.} Global BA is a special case of local BA, where all keyframes and map points are included in the optimization, except the origin keyframe, which is kept fixed to prevent gauge freedom. 

\subsection{Structured 3D Gaussians}
\label{sec:sup_s3d}

\begin{figure}[t]
	\centering
    \includegraphics[width=0.9\linewidth]{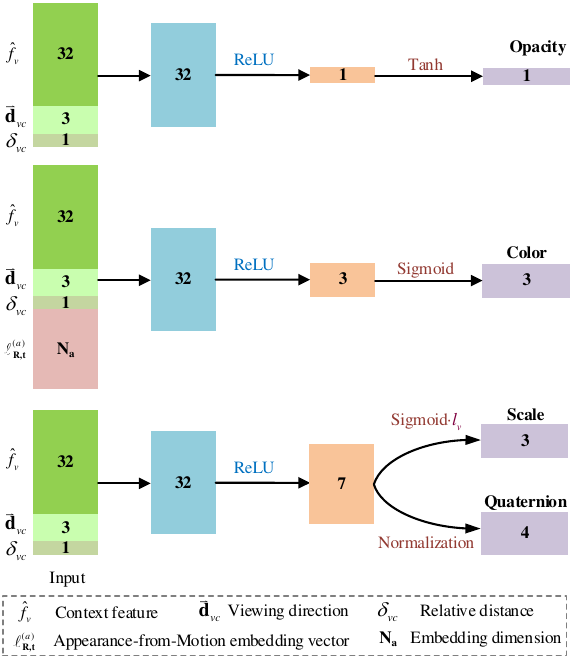}
\caption{\textbf{Structure of the MLPs} $M_\alpha$, $M_C$, $M_s$, and $M_q$. For each anchor, we use these MLPs to predict the opacity, color, scale, and quaternion of $k$ 3D Gaussian. The inputs to the MLPs include the relative distance $\delta_{vc}$ and the viewing direction  $\vec{\mathrm{\mathbf{d}}}_{vc}$ between camera position $\mathrm{\mathbf{t}}_c$ and an anchor point. Since $N_a$ is not a fixed parameter, its specific value is not included in the figure.}
\label{fig:MLP_anchor}
\end{figure}

\noindent\textbf{MLPs as feature decoders.}
Following \cite{Scaffold-GS2024}, we employ four MLPs as decoders to derive the parameters of each 3D Gaussian, including the opacity MLP $M_\alpha$, the color MLP $M_C$, and the covariance MLP $M_q, M_s$. Each MLP adopts a linear layer followed by ReLU and another linear layer. The outputs are activated by their respective activation functions to obtain the final parameters of each 3D Gaussian. The detailed architecture of these MLPs is illustrated in \cref{fig:MLP_anchor}. In our implementation, the hidden layer dimensions of all MLPs are set to 32.
\begin{itemize}
    \item For \textit{opacity}, we use $\mathrm{Tanh(\cdot)}$ to activate the output  of the final linear layer. Since the opacity values of 3D Gaussians are typically positive, we constrain the value range to $[0, 1)$ to ensure valid outputs.
    \item For \textit{color}, we use $\mathrm{Sigmoid}$ function to activate the output of the final linear layer, which constrains the color value into a range of $[0, 1)$.
    \item For \textit{rotation}, following 3D-GS \cite{3DGS2023}, we employ a normalization to activate the output of the final linear layer, ensuring the validity of the quaternion representation for rotation.
    \item For \textit{scaling}, a $\mathrm{Sigmoid}$ function is applied to activate the output of the final linear layer. Finally, the scaling of each 3D Gaussian is determined by adjusting the scaling $l_v$ of its associated anchor based on the MLP's output, as formulated below:
\begin{align}
\{s_0,\dots,s_{k-1}\} = M_s(\hat{f}_v,\ \delta_{vc},\ \vec{\mathrm{\mathbf{d}}}_{vc}) \cdot l_v
\label{eq:scaling_mlp1}
\end{align}
\end{itemize}

\subsection{Appearance-from-Motion Embedding}
\label{sec:sup_afme}

\begin{figure}[t]
	\centering
    \includegraphics[width=1\linewidth]{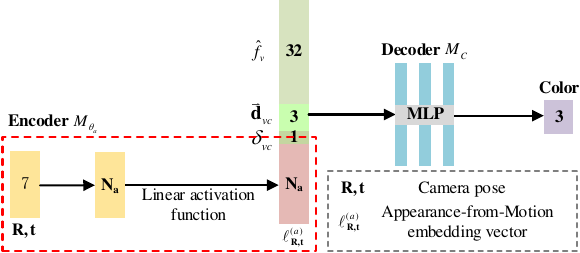}
\caption{\textbf{Structure of the MLPs} $M_{\theta_a}$ and $M_C$. We adopt an encoder-decoder architecture, where an encoder MLP $M_{\theta_a}$ first extracts features from the camera poses. For each anchor, the feature $\bm{\ell}^{(a)}_{\mathrm{\mathbf{R}},\ \mathrm{\mathbf{t}}}$, context feature $\hat{f}_v$, the relative distance $\delta_{vc}$ between camera position $\mathrm{\mathbf{t}}_c$ and the anchor point, and their viewing direction $\vec{\mathrm{\mathbf{d}}}_{vc}$ are then fed into a decoder $M_C$ to predict the color of each Gaussian.}
\label{fig:MLP_afme}
\end{figure}

\noindent\textbf{MLP as feature encoder.}
For AfME, we employ an MLP $M_{\theta_a}$ as the encoder. The input to this MLP is the camera pose corresponding to each image. The MLP $M_{\theta_a}$ extracts pose features and feeds these features to the color decoder $M_C$. The MLP $M_{\theta_a}$ adopts a structure of a linear layer followed by a linear activation function, as illustrated in \cref{fig:MLP_afme}. The entire pipeline for obtaining the Gaussian color is also detailed in \cref{fig:MLP_afme}. We adopt an encoder-decoder architecture, where an encoder MLP $M_{\theta_a}$ extracts features from the camera poses. Unlike FreGS \cite{FreGS2024}, we leverage only high-frequency information, as low-frequency components typically represent scene structure, which is already effectively captured by our structured 3D Gaussians.

\begin{table*}
\small
\setlength\tabcolsep{4.5pt}
  \centering
  \begin{tabular}{l|c|ccccccccc|cccc} 
    \toprule \noalign{\vskip -2pt}
    \multicolumn{2}{c|}{Datasets}  & \multicolumn{9}{c|}{Replica } & \multicolumn{4}{c}{TUM RGB-D}\\ 
    \midrule
    Method & Metric & R0 & R1 & R2 & Of0 & Of1 & Of2 & Of3 & Of4 & Avg. & fr1/d & fr2/x & fr3/o & Avg.\\
    \midrule
    \multirow{3}{*}{MonoGS \cite{MonoGS2024}} &  PSNR$\uparrow$   & \cellcolor{lightyellow}{34.29} & 35.77 & 36.79 & \cellcolor{lightyellow}{40.87} & 40.73 & \cellcolor{lightorange}{35.22} & \cellcolor{lightorange}{35.89} & 34.98 & 36.81 &\cellcolor{lightorange}{23.59}& \cellcolor{lightorange}{24.46} & \cellcolor{lightorange}{24.29} & \cellcolor{lightorange}{24.11}\\
    &SSIM$\uparrow$   & 0.953 & 0.957 & 0.965 & 0.979 & 0.977 & 0.961 & \cellcolor{lightyellow}{0.962} & 0.955 & 0.964 &\cellcolor{lightorange}{0.783} & \cellcolor{lightyellow}{0.789} & \cellcolor{lightorange}{0.829} & \cellcolor{lightorange}{0.800}\\
    &LPIPS$\downarrow$  & 0.071 & 0.078 & 0.074 & 0.048 & 0.052 & 0.074 & 0.061 & 0.092 & 0.069 &\cellcolor{lightyellow}{0.244} & 0.227 & 0.223 & 0.231\\
    
    \midrule
    \multirow{3}{*}{Photo-SLAM \cite{Photo-SLAM2024}} &  PSNR$\uparrow$   & 32.09 & 34.15 & 35.91 & 38.70 & 39.53 & 33.13 & 34.15 & 36.35 & 35.50 &20.14 & 22.15 &20.68 &20.99\\
    &SSIM$\uparrow$   & 0.920 & 0.941 & 0.959 & 0.967 & 0.964 & 0.943 & 0.943 & 0.956 & 0.949 &0.722& 0.765 &0.721 &0.736\\
    &LPIPS$\downarrow$  & 0.069 & 0.055 & \cellcolor{lightyellow}{0.041} & 0.048 & 0.045 & 0.075 & 0.064 & 0.053 & 0.056 &0.258 & \cellcolor{lightorange}{0.169} &\cellcolor{lightyellow}{0.211} &\cellcolor{lightyellow}{0.213}\\
    \midrule

    \multirow{3}{*}{Photo-SLAM-30K } &  PSNR$\uparrow$   & 31.41 & \cellcolor{lightyellow}{35.84} & \cellcolor{lightorange}{38.41} & 40.44 & 41.06 & \cellcolor{lightyellow}{34.56} & \cellcolor{lightyellow}{35.43} & \cellcolor{lightyellow}{38.36} & \cellcolor{lightyellow}{36.94}  & \cellcolor{lightyellow}{21.78} & 21.57 & \cellcolor{lightyellow}{21.84} & \cellcolor{lightyellow}{21.73}\\
    &SSIM$\uparrow$   & 0.873 & 0.955 & \cellcolor{lightyellow}{0.971} & 0.975 & 0.972 & 0.952 & 0.954 & 0.967 & 0.952 & \cellcolor{lightyellow}{0.766} & 0.755 & \cellcolor{lightyellow}{0.751} & 0.757\\
    &LPIPS$\downarrow$  & \cellcolor{lightorange}{0.046} & \cellcolor{lightorange}{0.036} & \cellcolor{lightorange}{0.026} & \cellcolor{lightyellow}{0.033} & \cellcolor{lightyellow}{0.033} & \cellcolor{lightyellow}{0.059} & \cellcolor{lightorange}{0.049} & \cellcolor{lightorange}{0.036} & \cellcolor{lightorange}{0.040} &\cellcolor{lightorange}{0.212}& \cellcolor{lightyellow}{0.182} &\cellcolor{lightorange}{0.165} & \cellcolor{lightorange}{0.186}\\
    
    \midrule
    \multirow{3}{*}{RTG-SLAM \cite{RTG-SLAM2024}} &  PSNR$\uparrow$   & 28.49 & 31.27 & 32.96 & 37.32 & 36.12 & 31.14 & 31.19 & 33.81 & 32.79 &13.62 & 17.08 & 18.70 & 16.47\\
    &SSIM$\uparrow$   & 0.834 & 0.902 & 0.927 & 0.957 & 0.943 & 0.923 & 0.918 & 0.937 & 0.918 &0.501 & 0.573 & 0.648 & 0.574\\
    &LPIPS$\downarrow$  & 0.152 & 0.119 & 0.122 & 0.084 & 0.103 & 0.145 & 0.139 & 0.125 & 0.124 &0.557 & 0.403 & 0.422 & 0.461\\
    
    \midrule
    \multirow{3}{*}{GS-SLAM$^*$ \cite{GS-SLAM2024}} &  PSNR$\uparrow$   & 31.56 & 32.86 &32.59 &38.70 &\cellcolor{lightyellow}{41.17} &32.36 &32.03 &32.92 &34.27 &- & - &- &-\\
    &SSIM$\uparrow$   & \cellcolor{lightorange}{0.968} & \cellcolor{lightorange}{0.973} & \cellcolor{lightyellow}{0.971} & \cellcolor{lightred}{\bf 0.986} & \cellcolor{lightred}{\bf 0.993} & \cellcolor{lightred}{\bf 0.978} & \cellcolor{lightred}{\bf 0.970} & \cellcolor{lightyellow}{0.968} & \cellcolor{lightred}{\bf 0.975} &- & - &- &-\\
    &LPIPS$\downarrow$  & 0.094 & 0.075 & 0.093 & 0.050 & \cellcolor{lightyellow}{0.033} & 0.094 & 0.110 & 0.112 & 0.082 &- & - &- &-\\
    
    \midrule
    \multirow{3}{*}{SplaTAM \cite{SplaTAM2024}} &  PSNR$\uparrow$   & 32.54 & 33.58 & 35.03 & 38.00 & 38.85 & 31.71 & 29.74 & 31.40 & 33.85 & 21.02 & \cellcolor{lightyellow}{23.39} & 19.81 & 21.41\\
    &SSIM$\uparrow$   & 0.938 & 0.936 & 0.952 & 0.963 & 0.955 & 0.928 & 0.902 & 0.914 & 0.936 & 0.753 & \cellcolor{lightorange}{0.806} & 0.731 & \cellcolor{lightyellow}{0.764}\\
    &LPIPS$\downarrow$  & 0.068 & 0.096 & 0.072 & 0.087 & 0.095 & 0.100 & 0.119 & 0.157 & 0.099 & 0.341 & 0.204 & 0.249 & 0.265\\ 
    
    \midrule
    \multirow{3}{*}{SGS-SLAM \cite{Sgs-slam2024}} &  PSNR$\uparrow$   & 32.48 & 33.50 & 35.11 & 38.22 & 38.91 & 31.86 & 30.05 & 31.53 & 33.96 &- & - &- &-\\
    &SSIM$\uparrow$   & \cellcolor{lightred}{\bf0.975} & \cellcolor{lightyellow}{0.968} & \cellcolor{lightred}{\bf 0.983} & \cellcolor{lightorange}{0.983} & \cellcolor{lightorange}{0.982} & 0.966 & 0.952 & 0.946 & \cellcolor{lightorange}{0.969}  &- & - &- &-\\
    &LPIPS$\downarrow$  & 0.071 & 0.099 & 0.073 & 0.083 & 0.091 & 0.099 & 0.118 & 0.154 & 0.099 &- & - &- &-\\ 
    
    \midrule
    \multirow{3}{*}{GS-ICP SLAM \cite{GS-ICPSLAM2024}}&  PSNR$\uparrow$   &\cellcolor{lightorange}{34.89}	&\cellcolor{lightorange}{37.15}	&\cellcolor{lightyellow}{37.89}	&\cellcolor{lightorange}{41.62}	&\cellcolor{lightorange}{42.86}	&32.69	&31.45	&\cellcolor{lightorange}{38.54}	&\cellcolor{lightorange}{37.14}  & 15.67 & 18.49 & 19.25 & 17.81\\
    &SSIM$\uparrow$    &\cellcolor{lightyellow}{0.955}	&0.965	&0.970	&0.981	&\cellcolor{lightyellow}{0.981}	&\cellcolor{lightyellow}{0.965}	&0.959	&\cellcolor{lightorange}{0.969}	&\cellcolor{lightyellow}{0.968} & 0.574 & 0.667 & 0.692 & 0.642\\
    &LPIPS$\downarrow$  &\cellcolor{lightyellow}{0.048}	&\cellcolor{lightyellow}{0.045}	&0.047	&\cellcolor{lightorange}{0.027}	&\cellcolor{lightorange}{0.031}	&\cellcolor{lightorange}{0.057}	&\cellcolor{lightyellow}{0.057}	&\cellcolor{lightyellow}{0.045}	&\cellcolor{lightyellow}{0.045} &0.444 & 0.308 & 0.329 & 0.361\\
    
    \midrule
    \multirow{3}{*}{\bf Ours} 
    &  PSNR$\uparrow$   & \cellcolor{lightred}{\bf37.07} & \cellcolor{lightred}{\bf39.54} & \cellcolor{lightred}{\bf40.33} & \cellcolor{lightred}{\bf 42.04} & \cellcolor{lightred}{\bf 43.21} & \cellcolor{lightred}{\bf36.38} & \cellcolor{lightred}{\bf37.18} & \cellcolor{lightred}{\bf39.62} & \cellcolor{lightred}{\bf39.42} & \cellcolor{lightred}{\bf 25.29} & \cellcolor{lightred}{\bf 26.35} & \cellcolor{lightred}{\bf 26.46} & \cellcolor{lightred}{\bf 26.03}\\
    &SSIM$\uparrow$  & \cellcolor{lightorange}{0.968} & \cellcolor{lightred}{\bf 0.977} & \cellcolor{lightorange}{0.980} & \cellcolor{lightyellow}{0.982} & 0.979 & \cellcolor{lightorange}{0.967} & \cellcolor{lightorange}{0.969} & \cellcolor{lightred}{\bf0.977} & \cellcolor{lightred}{\bf 0.975} & \cellcolor{lightred}{\bf 0.839} & \cellcolor{lightred}{\bf 0.831} & \cellcolor{lightred}{\bf 0.859} & \cellcolor{lightred}{\bf 0.843}\\
    &LPIPS$\downarrow$  & \cellcolor{lightred}{\bf0.023} & \cellcolor{lightred}{\bf0.016} & \cellcolor{lightred}{\bf0.015} & \cellcolor{lightred}{\bf0.020} & \cellcolor{lightred}{\bf0.019} & \cellcolor{lightred}{\bf0.035} & \cellcolor{lightred}{\bf0.026} & \cellcolor{lightred}{\bf0.018} & \cellcolor{lightred}{\bf0.021} & \cellcolor{lightred}{\bf 0.136} & \cellcolor{lightred}{\bf 0.081} & \cellcolor{lightred}{\bf 0.105} & \cellcolor{lightred}{\bf 0.107}\\
    \noalign{\vskip -2pt}\bottomrule
  \end{tabular}
  \caption{Quantitative evaluation of our method compared to state-of-the-art methods for {\bf RGB-D} camera on Replica and TUM RGB-D datasets. The best results are marked as  \colorbox{lightred}{\bf best score}, \colorbox{lightorange}{second best score} and \colorbox{lightyellow}{third best score}. GS-SLAM$^*$ denotes the result of GS-SLAM taken from \cite{GS-SLAM2024}, and all others are obtained in our experiments. '-' denotes that the system does not provide valid results. 
  }
  \label{tab:render_rgbd_sup}
\end{table*}

\begin{table*}[t]
\small
\setlength\tabcolsep{4.5pt}
  \centering
  \begin{tabular}{l|c|ccccccccc|cccc} 
    \toprule \noalign{\vskip -2pt}
    \multicolumn{2}{l|}{Datasets (Camera)}  & \multicolumn{9}{c|}{Replica ({\bf Mono}) } & \multicolumn{4}{c}{TUM RGB-D ({\bf Mono})}\\
    \midrule   
    Method & Metric & R0 & R1 & R2 & Of0 & Of1 & Of2 & Of3 & Of4 & Avg. & fr1/d & fr2/x & fr3/o & Avg. \\
    
    \midrule
    \multirow{3}{*}{MonoGS \cite{MonoGS2024}} 
    &  PSNR$\uparrow$   &26.19 & 25.42 & 27.83 & 31.90 &34.22 & 26.09 & 28.56 & 26.49 & 28.34 &20.38 & \cellcolor{lightorange}{21.21} & \cellcolor{lightorange}{21.41} & 21.00 \\
     &SSIM$\uparrow$   &0.819 & 0.798 & 0.889 & 0.911 & 0.930 & 0.881 & 0.898 & 0.897 & 0.878 & 0.691 & 0.690 & \cellcolor{lightorange}{0.735} & 0.705\\
     &LPIPS$\downarrow$  & 0.246 & 0.368 & 0.252 & 0.249 & 0.192 & 0.268 & 0.189 & 0.284 & 0.256 &0.377 & 0.377 & 0.426 & 0.393\\
    
    \midrule
    \multirow{3}{*}{Photo-SLAM \cite{Photo-SLAM2024}}
    &PSNR$\uparrow$   & 30.43 & 32.11 & 32.89 & 37.24 & 38.10 & 31.60 & 32.27 & 34.16 & 33.60 & 19.56 & 20.82 & 20.12 & 20.17\\
    &SSIM$\uparrow$   &0.890 & \cellcolor{lightorange}{0.926} & 0.937 & 0.960 & 0.955 & 0.932 & 0.928 & 0.943 & 0.934 & 0.705  & \cellcolor{lightorange}{0.718} & 0.702 &  0.708\\
    &LPIPS$\downarrow$  &0.099 &\cellcolor{lightorange}{0.073} &0.069 &0.062 &0.061 &0.094 &0.084 &0.073 &0.077 &0.281 & \cellcolor{lightorange}{0.158} & 0.233 & 0.224\\
    
    \midrule
    \multirow{3}{*}{Photo-SLAM-30K }
    &  PSNR$\uparrow$   & \cellcolor{lightorange}{32.13} & \cellcolor{lightorange}{33.14} & \cellcolor{lightorange}{37.27} & \cellcolor{lightorange}{38.04} & \cellcolor{lightorange}{41.73} & \cellcolor{lightorange}{35.22} & \cellcolor{lightorange}{34.88} & \cellcolor{lightorange}{36.22} & \cellcolor{lightorange}{36.08} & \cellcolor{lightorange}{22.57} & 20.54 & 20.08 & \cellcolor{lightorange}{21.06}\\
    &SSIM$\uparrow$   & \cellcolor{lightorange}{0.896} & 0.921 & \cellcolor{lightorange}{0.965} & \cellcolor{lightorange}{0.964} & \cellcolor{lightred}{\bf0.974} & \cellcolor{lightorange}{0.952} & \cellcolor{lightorange}{0.949} & \cellcolor{lightorange}{0.955} & \cellcolor{lightorange}{0.947} & \cellcolor{lightorange}{0.787} & 0.714 & 0.697 & \cellcolor{lightorange}{0.733}\\
    &LPIPS$\downarrow$  & \cellcolor{lightorange}{0.056} & 0.086 & \cellcolor{lightorange}{0.035} & \cellcolor{lightorange}{0.055} & \cellcolor{lightred}{\bf0.033} & \cellcolor{lightorange}{0.061} & \cellcolor{lightorange}{0.057} & \cellcolor{lightorange}{0.052}  & \cellcolor{lightorange}{0.054} & \cellcolor{lightorange}{0.179} & 0.166 & \cellcolor{lightorange}{0.213} & \cellcolor{lightorange}{ 0.186}\\
    
    \midrule
    \multirow{3}{*}{\bf Ours} 
    &  PSNR$\uparrow$   & \cellcolor{lightred}{\bf34.94} & \cellcolor{lightred}{\bf37.96} & \cellcolor{lightred}{\bf38.28} & \cellcolor{lightred}{\bf41.19} & \cellcolor{lightred}{\bf42.23} & \cellcolor{lightred}{\bf36.30} & \cellcolor{lightred}{\bf35.44} & \cellcolor{lightred}{\bf37.33} & \cellcolor{lightred}{\bf 37.96} &\cellcolor{lightred}{\bf 23.94} & \cellcolor{lightred}{\bf 25.39} &\cellcolor{lightred}{\bf 26.17} &\cellcolor{lightred}{\bf25.17}\\
    &SSIM$\uparrow$   & \cellcolor{lightred}{\bf0.949} & \cellcolor{lightred}{\bf0.967} & \cellcolor{lightred}{\bf0.971} & \cellcolor{lightred}{\bf0.978} & \cellcolor{lightorange}{0.972} & \cellcolor{lightred}{\bf0.964} & \cellcolor{lightred}{\bf0.952} & \cellcolor{lightred}{\bf 0.959} & \cellcolor{lightred}{\bf0.964}  &\cellcolor{lightred}{\bf 0.804} & \cellcolor{lightred}{\bf 0.813} &\cellcolor{lightred}{\bf 0.857} &\cellcolor{lightred}{\bf0.825}\\
    &LPIPS$\downarrow$  & \cellcolor{lightred}{\bf0.039} & \cellcolor{lightred}{\bf0.027} & \cellcolor{lightred}{\bf0.026} & \cellcolor{lightred}{\bf0.027} & \cellcolor{lightorange}{0.036} & \cellcolor{lightred}{\bf0.038} & \cellcolor{lightred}{\bf0.055} & \cellcolor{lightred}{\bf0.050} & \cellcolor{lightred}{\bf0.037} &\cellcolor{lightred}{\bf 0.135} & \cellcolor{lightred}{\bf 0.121} &\cellcolor{lightred}{\bf 0.110} &\cellcolor{lightred}{\bf0.122}\\
    \bottomrule
  \end{tabular}
  \caption{Quantitative evaluation of our method compared to state-of-the-art methods for {\bf Monocular (Mono)} camera on Replica and TUM RGB-D datasets. The best results are marked as   \colorbox{lightred}{\bf best score} and \colorbox{lightorange}{second best score}.}
  \label{tab:rendering_mono_sup}
\end{table*}

\begin{table*}
\small
\setlength\tabcolsep{2pt}
  \centering
  \begin{tabular}{l|c|ccccccccc|cccc} 
    \toprule \noalign{\vskip -2pt}
    \multicolumn{2}{c|}{Datasets}  & \multicolumn{9}{c|}{Replica } & \multicolumn{4}{c}{TUM RGB-D}\\ 
    \midrule
    Method & Metric & R0 & R1 & R2 & Of0 & Of1 & Of2 & Of3 & Of4 & Avg. & fr1/d & fr2/x & fr3/o & Avg.\\
    \midrule

    ORB-SLAM3 \cite{ORB-SLAM32021} 
    &RMSE$\downarrow$  & 0.500 & 0.537 & 0.731 & 0.762 & 1.338 & 0.636 & 0.419 & 9.319 & 1.780 & 5.056 & 0.390 & \cellcolor{lightyellow}{1.143} & 2.196\\

    DRIOD-SLAM \cite{DROID-SLAM2021} 
    &RMSE$\downarrow$  & 95.994 & 52.471 & 62.908 & 54.807 & 36.038 & 118.191 & 94.200 & 79.510 & 74.264 & 36.057 & 16.749 & 169.844 & 74.216\\

    MonoGS \cite{MonoGS2024} 
    &RMSE$\downarrow$  & 0.444 & 0.273 & 0.274 & 0.442 & 0.469 & \cellcolor{lightyellow}{0.220} & \cellcolor{lightorange}{0.159} & 2.237 & 0.565 & \cellcolor{lightred}{\bf1.531} & 1.440 & 1.535 & \cellcolor{lightorange}{1.502}\\

    Photo-SLAM \cite{Photo-SLAM2024}
    &RMSE$\downarrow$  & 0.529 & 0.397 & 0.295 & 0.501 & 0.379 & 1.202 & 0.768 & 0.585 & 0.582 & 3.578 & \cellcolor{lightred}{\bf0.337} & 1.696 & 1.870\\

    RTG-SLAM \cite{RTG-SLAM2024}
    &RMSE$\downarrow$  & \cellcolor{lightorange}{0.222} & 0.258 & \cellcolor{lightyellow}{0.248} & \cellcolor{lightred}{\bf0.201} & \cellcolor{lightred}{\bf0.190} & \cellcolor{lightred}{\bf0.115} & \cellcolor{lightred}{\bf0.156} & \cellcolor{lightorange}{0.136} & \cellcolor{lightorange}{0.191} & \cellcolor{lightorange}{1.582} & \cellcolor{lightyellow}{0.377} & \cellcolor{lightred}{\bf0.996} & \cellcolor{lightred}{\bf0.985}\\
    
    GS-SLAM$^*$ \cite{GS-SLAM2024}
    &RMSE$\downarrow$  & 0.480 & 0.530 & 0.330 & 0.520 & 0.410 & 0.590& 0.460 & 0.700 & 0.500 &3.300 & 1.300 & 6.600 &3.700\\

    SplaTAM \cite{SplaTAM2024}
    &RMSE$\downarrow$  & 0.501 & \cellcolor{lightyellow}{0.220} & 0.298 & \cellcolor{lightyellow}{0.316} & 0.582 & 0.256 & 0.288 & \cellcolor{lightyellow}{0.279} & \cellcolor{lightyellow}{0.343} & 5.102 & 1.339 & 3.329 & 4.215\\ 

    SGS-SLAM \cite{Sgs-slam2024}
    &RMSE$\downarrow$  & 0.463 & \cellcolor{lightorange}{0.216} & 0.300 & \cellcolor{lightorange}{0.339} & 0.547 & 0.299 & 0.451 & 0.311 & 0.365 &- & - &- &-\\ 

    GS-ICP SLAM \cite{GS-ICPSLAM2024}
    &RMSE$\downarrow$  & \cellcolor{lightred}{\bf0.189} & \cellcolor{lightred}{\bf0.132} & \cellcolor{lightorange}{0.216} & \cellcolor{lightred}{\bf0.201} & \cellcolor{lightorange}{0.236} & \cellcolor{lightorange}{0.160} & \cellcolor{lightyellow}{0.162} & \cellcolor{lightred}{\bf0.117}	&\cellcolor{lightred}{\bf0.177} & 3.539 & 2.251 & 2.972 & 2.921\\

    {\bf Ours} 
    &RMSE$\downarrow$  & \cellcolor{lightyellow}{0.296} & 0.264 & \cellcolor{lightred}{\bf 0.182} & 0.429 & \cellcolor{lightyellow}{0.354} & 1.040 & 0.434 & 0.441 & 0.430 & \cellcolor{lightyellow}{3.187} & \cellcolor{lightorange}{0.370} & \cellcolor{lightorange}{1.026} &\cellcolor{lightyellow}{1.528}\\
    \noalign{\vskip -2pt}\bottomrule
  \end{tabular}
  \caption{Camera tracking result on Replica and TUM RGB-D datasets for \textbf{RGB-D} camera.  {\bf RMSE of ATE (\rm{cm})} is reported. The best results are marked as   \colorbox{lightred}{\bf best score}, \colorbox{lightorange}{second best score} and \colorbox{lightyellow}{third best score}.  '-' denotes the system does not provide valid results.}
  \label{tab:ate_rgbd_sup}
\end{table*}

\begin{table*}[t]
\small
\setlength\tabcolsep{1.25pt}
  \centering
  \begin{tabular}{l|c|ccccccccc|cccc} 
    \toprule \noalign{\vskip -2pt}
    \multicolumn{2}{l|}{Datasets (Camera)}  & \multicolumn{9}{c|}{Replica ({\bf Mono}) } & \multicolumn{4}{c}{TUM RGB-D ({\bf Mono})}\\
    \midrule   
    Method & Metric & R0 & R1 & R2 & Of0 & Of1 & Of2 & Of3 & Of4 & Avg. & fr1/d & fr2/x & fr3/o & Avg. \\
    \midrule

    ORB-SLAM3 \cite{ORB-SLAM32021} 
    &RMSE$\downarrow$  & 51.388 & \cellcolor{lightyellow}{26.384} & \cellcolor{lightyellow}{4.330} & 110.212 & 103.948 & 65.359 & 51.145 & \cellcolor{lightyellow}{1.188} & 51.744 & 4.3269 & 10.4598 & 123.226 & 46.004\\

    DRIOD-SLAM \cite{DROID-SLAM2021} 
    &RMSE$\downarrow$  & 103.892 & 53.146 & 66.939 & 53.267 & 34.431 & 119.311 & 98.089 & 83.732 & 76.600 & \cellcolor{lightorange}{1.769} & \cellcolor{lightorange}{0.458} & \cellcolor{lightyellow}{2.839} & \cellcolor{lightyellow}{1.689}\\

    MonoGS \cite{MonoGS2024} 
    &RMSE$\downarrow$  & \cellcolor{lightyellow}{12.623} & 56.357 & 25.350 & \cellcolor{lightyellow}{43.245} & \cellcolor{lightyellow}{19.729} & \cellcolor{lightyellow}{39.148} & \cellcolor{lightyellow}{11.754} & 88.230 & \cellcolor{lightyellow}{37.054} & 4.575 & 4.605 & 2.847 & 4.009\\

    Photo-SLAM \cite{Photo-SLAM2024}
    &RMSE$\downarrow$  & \cellcolor{lightorange}{0.336} & \cellcolor{lightorange}{0.551} & \cellcolor{lightorange}{0.234} & \cellcolor{lightorange}{2.703} & \cellcolor{lightorange}{0.505} & \cellcolor{lightred}{\bf2.065} & \cellcolor{lightorange}{0.399} & \cellcolor{lightorange}{0.644} & \cellcolor{lightorange}{0.930} & \cellcolor{lightred}{\bf1.633} & \cellcolor{lightyellow}{0.935} & \cellcolor{lightorange}{2.050} &\cellcolor{lightorange}{ 1.539}\\

    {\bf Ours} 
    &RMSE$\downarrow$  & \cellcolor{lightred}{\bf0.288} & \cellcolor{lightred}{\bf0.388} & \cellcolor{lightred}{\bf0.215} & \cellcolor{lightred}{\bf0.579} & \cellcolor{lightred}{\bf0.320} & \cellcolor{lightorange}{3.963} & \cellcolor{lightred}{\bf0.307} & \cellcolor{lightred}{\bf0.603} & \cellcolor{lightred}{\bf0.833} &\cellcolor{lightyellow}{3.187} & \cellcolor{lightred}{\bf0.370} & \cellcolor{lightred}{\bf1.026} &\cellcolor{lightred}{\bf1.505}\\
    \bottomrule
  \end{tabular}
  \caption{Camera tracking result on Replica and TUM RGB-D datasets for \textbf{monocular} camera.  {\bf RMSE of ATE (\rm{cm})} is reported. Best results are marked as   \colorbox{lightred}{\bf best score}, \colorbox{lightorange}{second best score} and \colorbox{lightyellow}{third best score}.}
  \label{tab:ate_mono_sup}
\end{table*}

\subsection{Anchor Points Refinement}
\label{sec:sup_anchor}
Our anchor refinement strategy follows \cite{Scaffold-GS2024}, and it is included here to enhance the completeness of this paper.

\noindent\textbf{Anchor Growing.}
3D Gaussians are spatially quantized into voxels of size $\epsilon_g = 0.001$. For all 3D Gaussians within each voxel, we compute the average gradient after $N_t =100$  training iterations, denoted as $\nabla_g $. When the average gradient $\nabla_g $ within a voxel exceeds a threshold $ \tau_g = 0.0002$, a new anchor is added at the center of the voxel. Since the total loss includes the frequency regularization $\mathcal{L}_{hf}$ in \cref{eq:loss_fre}, anchor points grow toward underrepresented high-frequency regions in the scene. Ultimately, the local details of the scene are refined. 
In our implementation, the scene is quantized into a multi-resolution voxel grid, allowing new anchors to be added to regions of varying sizes, as defined by
\begin{align}
\epsilon_g^{(m)} = \epsilon_g/4^{m-1}, \ \tau_g^{(m)} = \tau_g \cdot 2^{m-1}
\label{eq:anchor_growing}
\end{align}
where $m$ denotes the level of quantization. Additionally, we adopt a random candidate pruning strategy to moderate the growth rate of anchors.


To eliminate redundant anchor points, we evaluate their opacity. Specifically, after $N_t$ training iterations, we accumulate the opacity values of each 3D Gaussian. If the accumulated value $\alpha_{p}$ falls below a pre-defined threshold, the Gaussian is removed from the scene.

\subsection{Experimental Parameters}
For the monocular camera in the Replica dataset, the dimension of AfME $N_a$  is set to 1, while for other configurations, it is set to $32$. Each anchor manages $k = 10$ 3D Gaussians. Anchors with an opacity value below $0.005$ are removed. The loss weights  $\lambda$, $\lambda_{\text{vol}}$, and $\lambda_{hf}$ are set to $0.2$, $0.01$, and $0.01$, respectively. For the monocular camera in the Replica dataset, the weight $\lambda_{hf}$ for frequency regularization is set to $0.025$, and the frequency pyramid consists of $3$ levels.

\section{Additional Qualitative Results}
\label{sec:sup_result}

\subsection{Per-scene Results.}
\begin{figure*}[ht]
	\centering
    \begin{subfigure}[t]{1\linewidth}
        \includegraphics[width=1\linewidth]{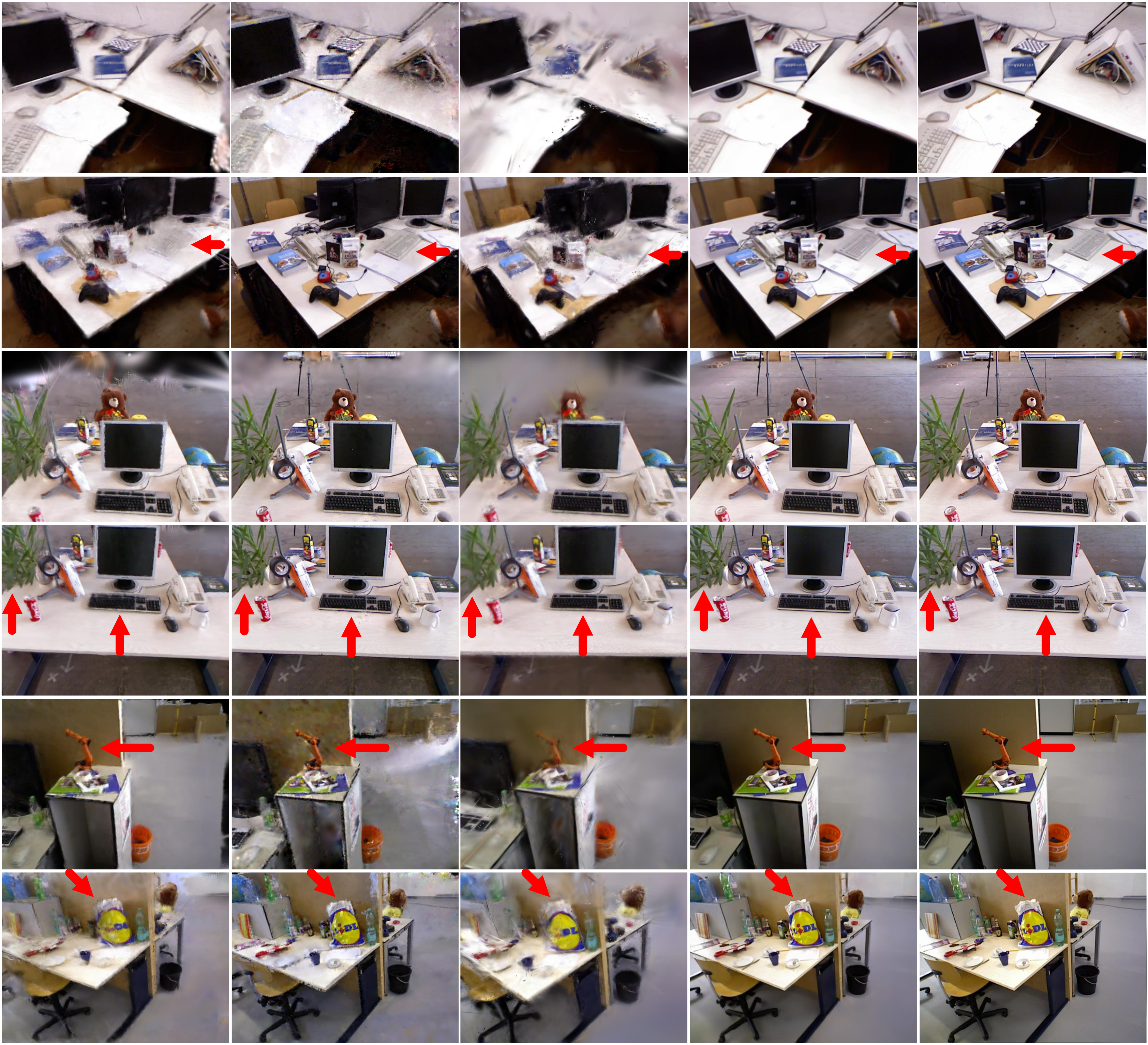}
    \end{subfigure}
    \begin{subfigure}[t]{0.19\linewidth}
        \vspace{-0.5cm}
        \caption{GS-ICP SLAM \cite{GS-ICPSLAM2024}}
    \end{subfigure}
    \begin{subfigure}[t]{0.19\linewidth}
        \vspace{-0.5cm}
        \caption{SplaTAM \cite{SplaTAM2024}}
    \end{subfigure}
    \begin{subfigure}[t]{0.19\linewidth}
        \vspace{-0.5cm}
        \caption{RTG-SLAM \cite{RTG-SLAM2024}}
    \end{subfigure}
    \begin{subfigure}[t]{0.19\linewidth}
        \vspace{-0.5cm}
        \caption{{\bf Ours}}
    \end{subfigure}
    \begin{subfigure}[t]{0.19\linewidth}
        \vspace{-0.5cm}
        \caption{Ground Truth}
    \end{subfigure}
	\caption{We show comparisons of ours to state-of-the-art methods on TUM RGB-D dataset for \textbf{RGB-D} camera. From top to bottom, the scenes are \textit{fr1/desk} (rows 1–2), \textit{fr2/xyz} (rows 3–4), and \textit{fr3/office} (rows 5–6). Non-obvious differences in quality are highlighted by arrows.
    }
	\label{fig:rendering_tumrgbd_supple}
\end{figure*}

\begin{figure*}[ht]
	\centering
    \begin{subfigure}[t]{1\linewidth}
        \includegraphics[width=1\linewidth]{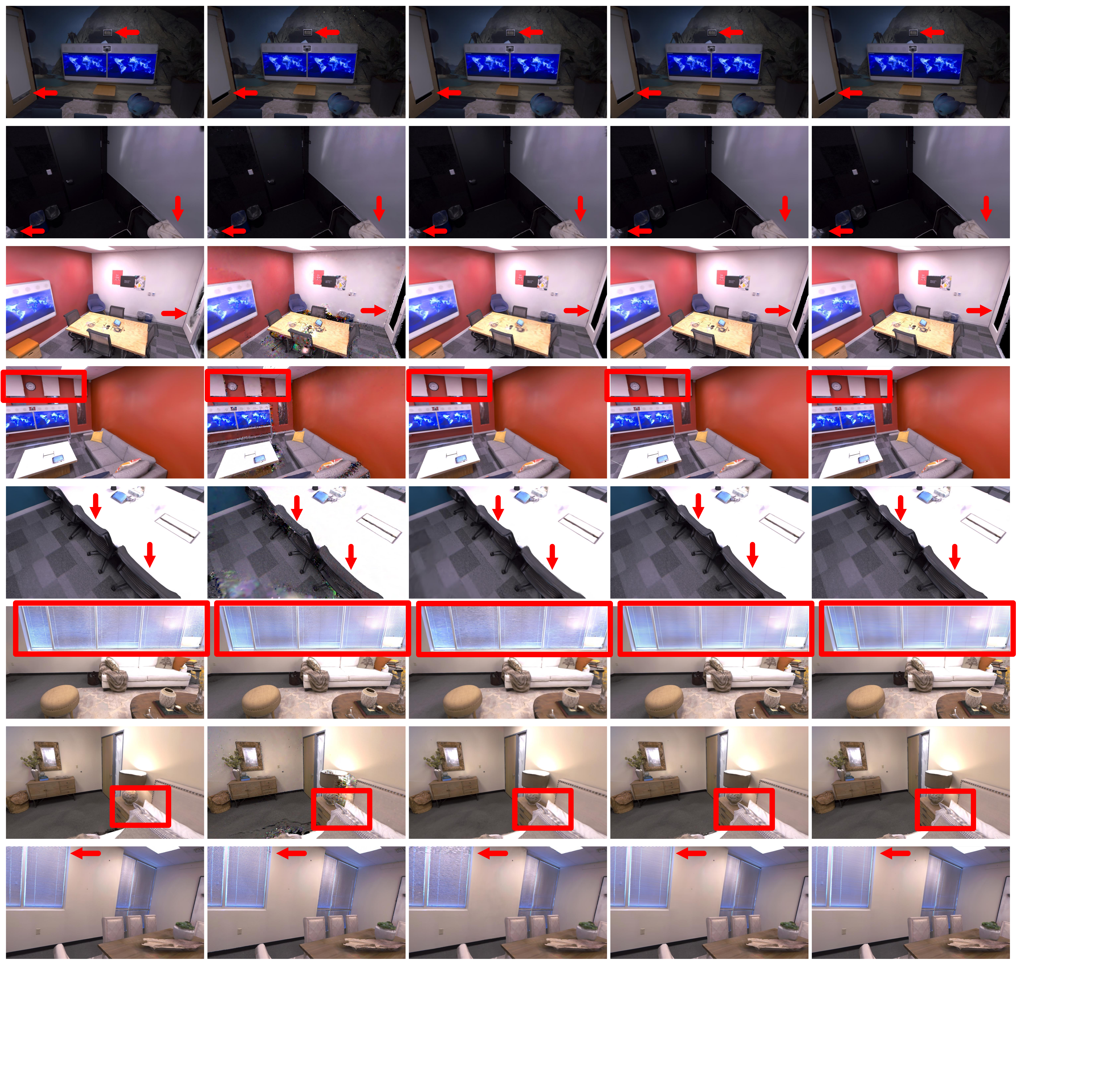}
    \end{subfigure}
    \begin{subfigure}[t]{0.19\linewidth}
        \vspace{-0.5cm}
        \caption{GS-ICP SLAM \cite{GS-ICPSLAM2024}}
    \end{subfigure}
    \begin{subfigure}[t]{0.19\linewidth}
        \vspace{-0.5cm}
        \caption{SplaTAM \cite{SplaTAM2024}}
    \end{subfigure}
    \begin{subfigure}[t]{0.19\linewidth}
        \vspace{-0.5cm}
        \caption{RTG-SLAM \cite{RTG-SLAM2024}}
    \end{subfigure}
    \begin{subfigure}[t]{0.19\linewidth}
        \vspace{-0.5cm}
        \caption{{\bf Ours}}
    \end{subfigure}
    \begin{subfigure}[t]{0.19\linewidth}
        \vspace{-0.5cm}
        \caption{Ground Truth}
    \end{subfigure}
	\caption{We show comparisons of ours to state-of-the-art methods on Replica dataset for \textbf{RGB-D} camera. From top to bottom, the scenes are \textit{Office0}, \textit{Office1}, \textit{Office2}, \textit{Office3}, \textit{Office4}, \textit{room0},  \textit{room1}, and \textit{room2}. Non-obvious differences in quality are highlighted by arrows/insets. Since all methods achieve high-quality rendering results on the Replica dataset, we have highlighted specific regions in the images. In these annotated areas, our method consistently demonstrates sharper edges or finer textures.
    }
	\label{fig:rendering_replicargbd_supple}
\end{figure*}

\begin{figure*}[ht]
	\centering
    \begin{subfigure}[t]{1\linewidth}
        \includegraphics[width=1\linewidth]{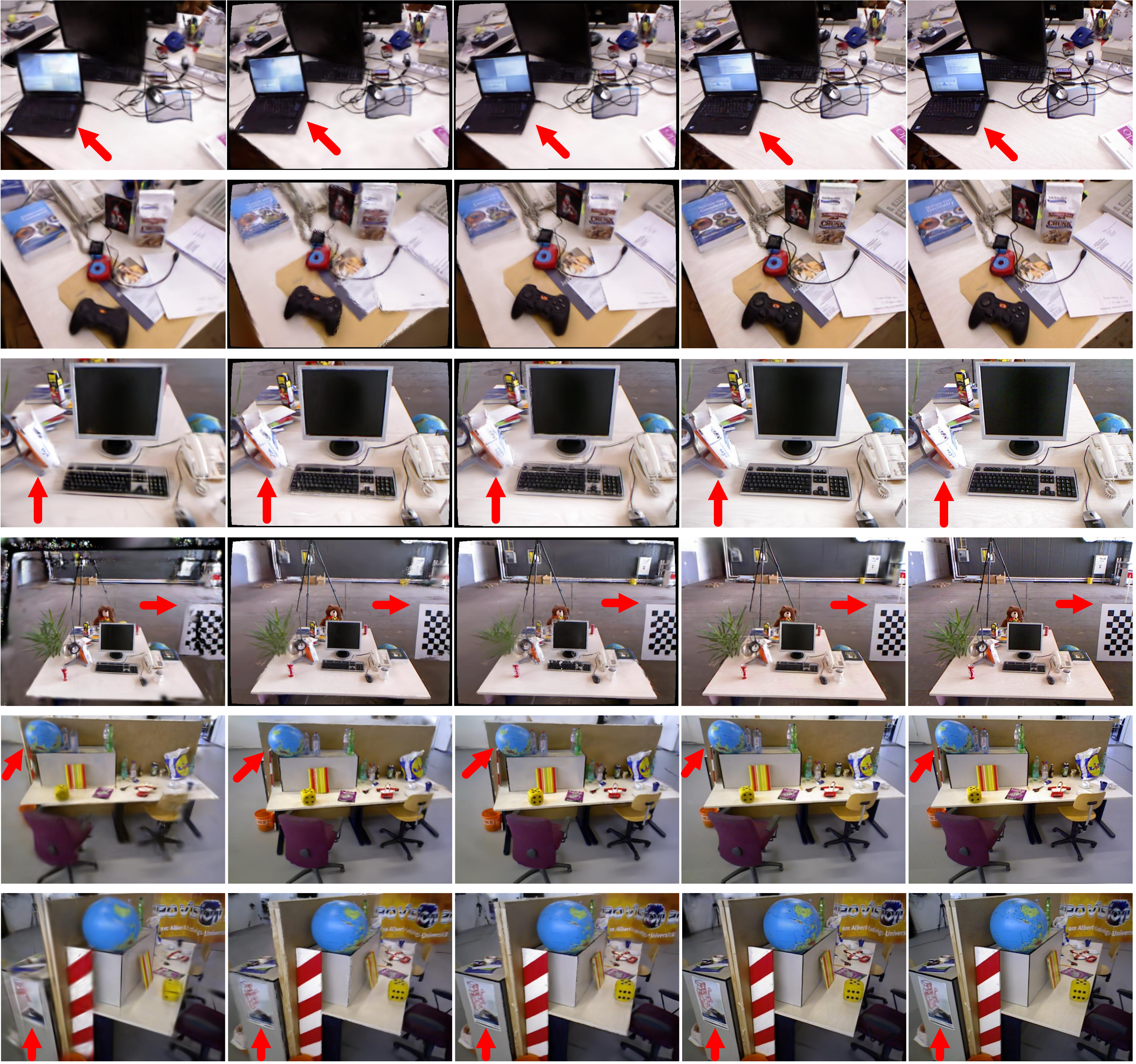}
    \end{subfigure}
    \begin{subfigure}[ht]{0.19\linewidth}
           \vspace{-0.15cm}
        \caption{MonoGS \cite{MonoGS2024}}
    \end{subfigure}
    \begin{subfigure}[ht]{0.19\linewidth}
            \vspace{-0.15cm}
        \caption{Photo-SLAM \cite{Photo-SLAM2024}}
    \end{subfigure}
    \begin{subfigure}[ht]{0.19\linewidth}
            \vspace{-0.15cm}
        \caption{Photo-SLAM-30K}
    \end{subfigure}
    \begin{subfigure}[ht]{0.19\linewidth}
            \vspace{-0.15cm}
        \caption{{\bf Ours}}
    \end{subfigure}
    \begin{subfigure}[ht]{0.19\linewidth}
           \vspace{-0.15cm}
        \caption{Ground Truth}
    \end{subfigure}
	\caption{We show comparisons of ours to state-of-the-art methods on TUM RGB-D dataset for \textbf{Monocular} camera. From top to bottom, the scenes are \textit{fr1/desk} (rows 1–2), \textit{fr2/xyz} (rows 3–4), and \textit{fr3/office} (rows 5–6). Non-obvious differences in qualityare  highlighted by arrows. Photo-SLAM \cite{Photo-SLAM2024} uses a set of parameters to undistort images as ground truth supervision. Consequently, its rendered images for \textit{fr1/desk} and \textit{fr2/xyz} exhibit black borders.}
	\label{fig:rendering_tumrmono_supple}
\end{figure*}

\begin{figure*}[ht]
	\centering
    \begin{subfigure}[t]{1\linewidth}
        \includegraphics[width=1\linewidth]{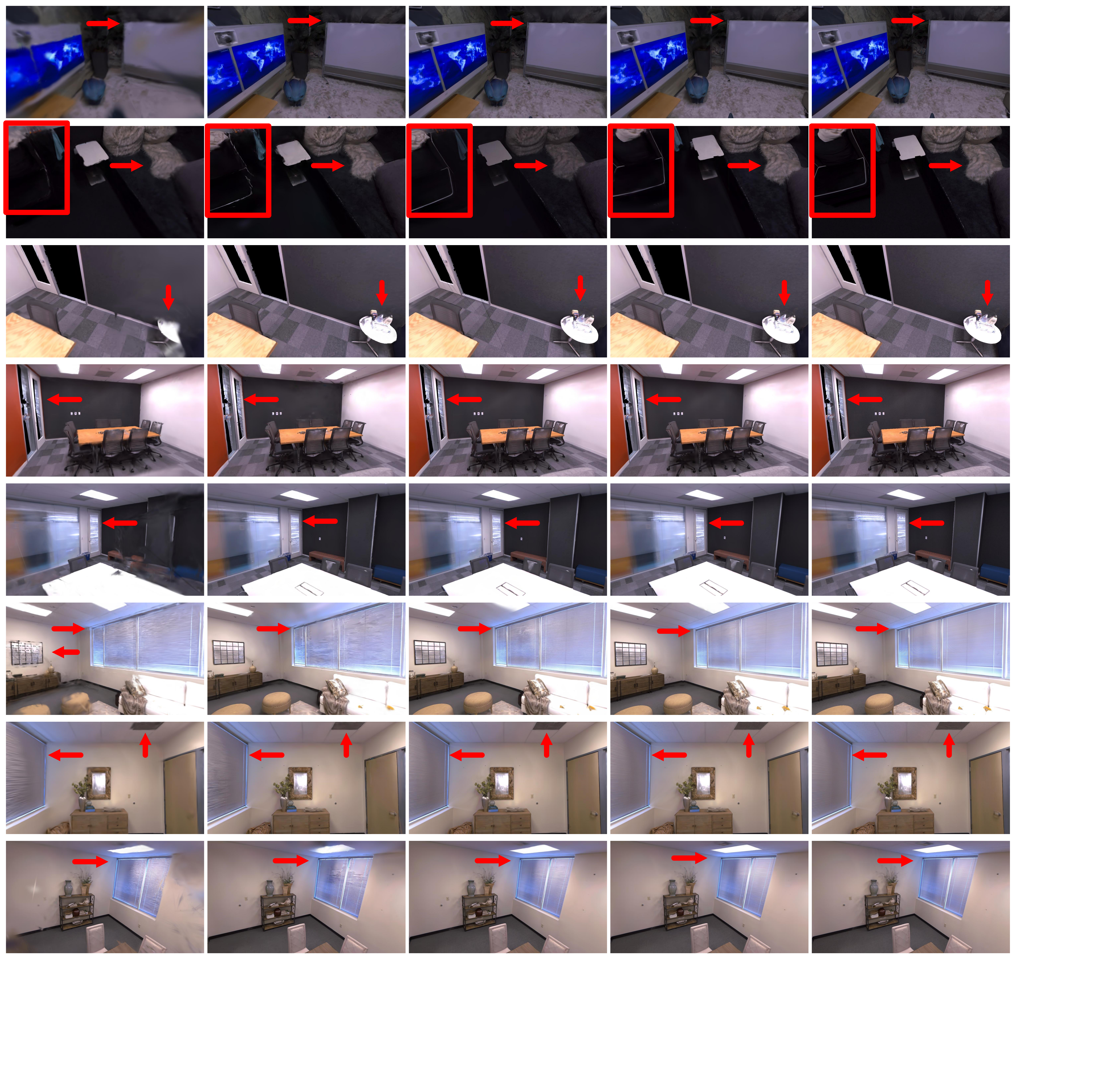}
    \end{subfigure}
    \begin{subfigure}[ht]{0.19\linewidth}
           \vspace{-0.15cm}
        \caption{MonoGS \cite{MonoGS2024}}
    \end{subfigure}
    \begin{subfigure}[ht]{0.19\linewidth}
            \vspace{-0.15cm}
        \caption{Photo-SLAM \cite{Photo-SLAM2024}}
    \end{subfigure}
    \begin{subfigure}[ht]{0.19\linewidth}
            \vspace{-0.15cm}
        \caption{Photo-SLAM-30K}
    \end{subfigure}
    \begin{subfigure}[ht]{0.19\linewidth}
            \vspace{-0.15cm}
        \caption{{\bf Ours}}
    \end{subfigure}
    \begin{subfigure}[ht]{0.19\linewidth}
           \vspace{-0.15cm}
        \caption{Ground Truth}
    \end{subfigure}
	\caption{We show comparisons of ours to state-of-the-art methods on Replica dataset for \textbf{Monocular} camera. From top to bottom, the scenes are \textit{Office0}, \textit{Office1}, \textit{Office2}, \textit{Office3}, \textit{Office4}, \textit{room0},  \textit{room1}, and \textit{room2}. Non-obvious differences in quality are highlighted by arrows/insets. Since all methods achieve high-quality rendering results on the Replica dataset, we have highlighted specific regions in the images. In these annotated areas, our method consistently demonstrates sharper edges or finer textures.}
	\label{fig:rendering_replicarmono_supple}
\end{figure*}

\begin{figure*}[ht]
	\centering
    \begin{subfigure}[t]{1\linewidth}
        \includegraphics[width=1\linewidth]{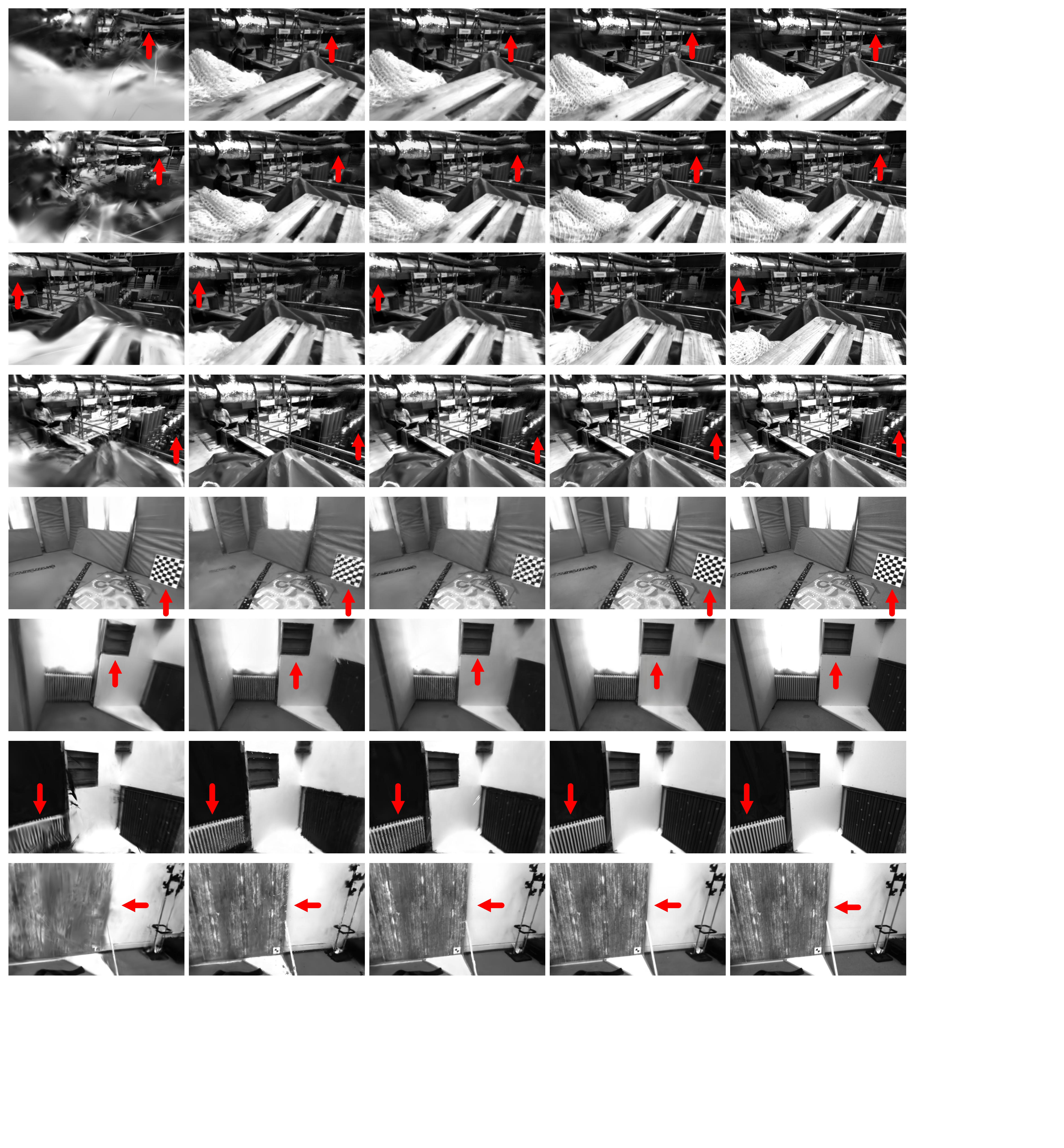}
    \end{subfigure}
    \begin{subfigure}[ht]{0.19\linewidth}
           \vspace{-0.15cm}
        \caption{MonoGS \cite{MonoGS2024}}
    \end{subfigure}
    \begin{subfigure}[ht]{0.19\linewidth}
            \vspace{-0.15cm}
        \caption{Photo-SLAM \cite{Photo-SLAM2024}}
    \end{subfigure}
    \begin{subfigure}[ht]{0.19\linewidth}
            \vspace{-0.15cm}
        \caption{Photo-SLAM-30K}
    \end{subfigure}
    \begin{subfigure}[ht]{0.19\linewidth}
            \vspace{-0.15cm}
        \caption{{\bf Ours}}
    \end{subfigure}
    \begin{subfigure}[ht]{0.19\linewidth}
           \vspace{-0.15cm}
        \caption{Ground Truth}
    \end{subfigure}
	\caption{
    We show comparisons of ours to state-of-the-art methods on the EuRoC MAV dataset for \textbf{Stereo} camera. From top to bottom, the scenes are \textit{MH01} (rows 1–2), \textit{MH02} (rows 3–4), \textit{V101} (rows 5–6), and \textit{V201} (rows 7–8). Non-obvious differences in quality are highlighted by arrows.
 }
	\label{fig:rendering_euorcstereo_supple}
\end{figure*}

\cref{tab:render_rgbd_sup}, \cref{tab:rendering_mono_sup}, \cref{tab:rendering_stereo_sup},
\cref{tab:ate_rgbd_sup},
\cref{tab:ate_mono_sup}, and
\cref{tab:ate_stereo_sup}
present the photorealistic mapping and localization results of our method across all datasets for each scene. Additionally, 
\cref{fig:rendering_tumrgbd_supple}, \cref{fig:rendering_replicargbd_supple}, 
\cref{fig:rendering_tumrmono_supple},
\cref{fig:rendering_replicarmono_supple},  and \cref{fig:rendering_euorcstereo_supple} show more rendering comparisons between our method and all baseline methods for each scene.



\begin{table*}[tb]
\small
  \centering
  \begin{subtable}[t]{0.49\textwidth} 
  \centering
  \begingroup
  \setlength\tabcolsep{4.5pt}
  \begin{tabular}{l|c|ccccc} 
    \toprule \noalign{\vskip -2pt}
    \multicolumn{2}{l|}{Datasets (Camera)}  & \multicolumn{5}{c}{EuRoC ({\bf Stereo})}\\
    
    \midrule   
    Method & Metric & MH01 & MH02 & V101 & V201 & Avg. \\
    
    \midrule
    MonoGS 
    &  PSNR$\uparrow$  & \cellcolor{lightred}{\bf22.84} & \cellcolor{lightred}{\bf25.53} & \cellcolor{lightorange}{23.39} & \cellcolor{lightorange}{18.66} & \cellcolor{lightorange}{22.60} \\
     \cite{MonoGS2024} &SSIM$\uparrow$   & \cellcolor{lightred}{\bf0.789} & \cellcolor{lightred}{\bf0.850} & \cellcolor{lightorange}{0.831} & \cellcolor{lightorange}{0.687} & \cellcolor{lightorange}{0.789} \\
     &LPIPS$\downarrow$  & \cellcolor{lightorange}{0.243} & \cellcolor{lightred}{\bf0.181} & \cellcolor{lightorange}{0.287} & \cellcolor{lightorange}{0.384} & \cellcolor{lightorange}{0.274}\\
    
    \midrule
    Photo-
    &PSNR$\uparrow$   & 11.22 & 11.14 & 13.78 & 11.46 & 11.90\\
    SLAM &SSIM$\uparrow$   & 0.300 & 0.306 & 0.520 & 0.509 & 0.409\\
    \cite{Photo-SLAM2024} &LPIPS$\downarrow$  & 0.469 & 0.464 & 0.394 & 0.427 & 0.439 \\
    
    \midrule
    Photo-
    &PSNR$\uparrow$  & 11.10 & 11.04 & 13.66 & 11.26 & 11.77 \\
    SLAM- &SSIM$\uparrow$   & 0.296 & 0.300 & 0.516 & 0.508 & 0.405 \\
    30K &LPIPS$\downarrow$  & 0.466 & 0.457 & 0.389 & 0.409 & 0.430 \\
    
    \midrule
    \multirow{3}{*}{\bf Ours} 
    &  PSNR$\uparrow$  & \cellcolor{lightorange}{22.50} & \cellcolor{lightorange}{22.30} & \cellcolor{lightred}{\bf24.90} & \cellcolor{lightred}{\bf24.89} & \cellcolor{lightred}{\bf23.64}\\
    &SSIM$\uparrow$    & \cellcolor{lightorange}{0.750} & \cellcolor{lightorange}{0.727} & \cellcolor{lightred}{\bf0.843} & \cellcolor{lightred}{\bf0.842} & \cellcolor{lightred}{\bf0.791} \\
    &LPIPS$\downarrow$  & \cellcolor{lightred}{\bf0.220} & \cellcolor{lightorange}{0.269} & \cellcolor{lightred}{\bf0.122} & \cellcolor{lightred}{\bf0.117} & \cellcolor{lightred}{\bf0.182}\\
    \bottomrule
  \end{tabular}
   \endgroup
  \caption{Quantitative evaluation of our method compared to state-of-the-art methods for {\bf Stereo} camera on EuRoC MAV datasets. The best results are marked as  \colorbox{lightred}{\bf best score} and \colorbox{lightorange}{second best score}.}
  \label{tab:rendering_stereo_sup}
  \end{subtable}
    \begin{subtable}[t]{0.49\textwidth} 
  \centering
  \begingroup
  \setlength\tabcolsep{0.75pt}
  \begin{tabular}{l|c|ccccc} 
    \toprule \noalign{\vskip -2pt}
    \multicolumn{2}{l|}{Datasets (Camera)}  & \multicolumn{4}{c}{EuRoC ({\bf Stereo})}\\
    
    \midrule   
    Method & Metric & MH01 & MH02 & V101 & V201 & Avg. \\
    \midrule
    ORB-SLAM3 \cite{ORB-SLAM32021} 
    &RMSE$\downarrow$   & 4.806 & 4.938 & 8.829 & 25.057 & 10.907\\

    DRIOD-SLAM \cite{DROID-SLAM2021} 
    &RMSE$\downarrow$   & \cellcolor{lightred}{\bf1.177} &\cellcolor{lightred}{\bf 1.169} & \cellcolor{lightred}{\bf3.678} & \cellcolor{lightred}{\bf1.680} & \cellcolor{lightred}{\bf1.926}\\

    MonoGS 
     \cite{MonoGS2024}
     &RMSE$\downarrow$  & 11.194 & 8.327 & 29.365 & 148.080 & 49.241 \\

    Photo-SLAM \cite{Photo-SLAM2024} 
    &RMSE$\downarrow$  & 3.997 & 4.547 & 8.882 & 26.665 & 11.023 \\

    {\bf Ours}
    &RMSE$\downarrow$  & \cellcolor{lightorange}{3.948 }& \cellcolor{lightorange}{3.863} & \cellcolor{lightorange}{8.823} & \cellcolor{lightorange}{13.217} & \cellcolor{lightorange}{7.462}\\
    \bottomrule
  \end{tabular}
   \endgroup
  \caption{Camera tracking result on EuRoC MAV datasets for \textbf{stereo} camera.  {\bf RMSE of ATE (\rm{cm})} is reported.  The best results are marked as  \colorbox{lightred}{\bf best score} and \colorbox{lightorange}{second best score}.}
  \label{tab:ate_stereo_sup}
    \end{subtable}
  \caption{Quantitative evaluation of our method compared to state-of-the-art methods for {\bf Stereo} camera on EuRoC MAV datasets.}
\end{table*}


\end{document}